\documentclass[letterpaper, 10 pt, conference]{ieeeconf}  

\IEEEoverridecommandlockouts                              

\usepackage[utf8]{inputenc}
\usepackage[T1]{fontenc}

\usepackage{graphicx}
\usepackage[table]{xcolor}
\usepackage{capt-of}
\usepackage{tikz}
\usepackage{comment}
\usepackage{amsmath,amssymb} 
\usepackage{color}
\usepackage{booktabs}
\usepackage{wrapfig}
\usepackage{xspace}
\definecolor{grey}{rgb}{0.95,0.95,0.95}
\usepackage[noadjust]{cite}

\usepackage[pagebackref,breaklinks,colorlinks]{hyperref}

\newcommand{\bx}{\mathbf{x}}

\newcommand{\bc}{\mathbf{c}}
\newcommand{\bd}{\mathbf{d}}

\newcommand{\bo}{\mathbf{o}}
\newcommand{\cL}{\mathcal{L}}

\def\eg{\emph{e.g}.} 
\def\ie{\emph{i.e}.} 
 
\def\etc{\emph{etc}.} 
\def\wrt{w.r.t.}

\def \short {} 
\ifx \short \undefined
   \newcommand{\cutsectionup}{}
   \newcommand{\cutsectiondown}{}

   \newcommand{\cutsubsectionup}{}
   \newcommand{\cutsubsectiondown}{}

   \newcommand{\cuthalfcaptionup}{}
   \newcommand{\cuthalfcaptiondown}{}

   \newcommand{\cuthalftablecaptionup}{}
   \newcommand{\cuthalftablecaptiondown}{}
\else
   \newcommand{\cutsectionup}{\vspace*{-3pt}}
   \newcommand{\cutsectiondown}{\vspace*{-1pt}}

   \newcommand{\cutsubsectionup}{\vspace*{-2pt}}
   \newcommand{\cutsubsectiondown}{\vspace*{-1pt}}

   \newcommand{\cuthalfcaptionup}{\vspace*{-12pt}}
   \newcommand{\cuthalfcaptiondown}{\vspace*{-8pt}}

   \newcommand{\cuthalftablecaptionup}{\vspace*{-10pt}}
   \newcommand{\cuthalftablecaptiondown}{\vspace*{-10pt}}
\fi

\def\name{NeuSim\xspace}

\title{\LARGE \bf
Reconstructing Objects in-the-wild for Realistic Sensor Simulation
}

\author{
Ze Yang$^{1,2}$, Sivabalan Manivasagam$^{1,2}$, Yun Chen$^{1,2}$, Jingkang Wang$^{1,2}$, Rui Hu$^{1}$, Raquel Urtasun$^{1,2}$ \\
Waabi$^{1}$, University of Toronto$^{2}$ \\
\texttt {\{zyang, siva, ychen, jwang, rhu, urtasun\}@waabi.ai}
}

\begin{document}

\twocolumn[{
\renewcommand\twocolumn[1][]{#1}
\maketitle
    \begin{center}
    \vspace{-5.0mm}
	\includegraphics[width=0.95\textwidth]{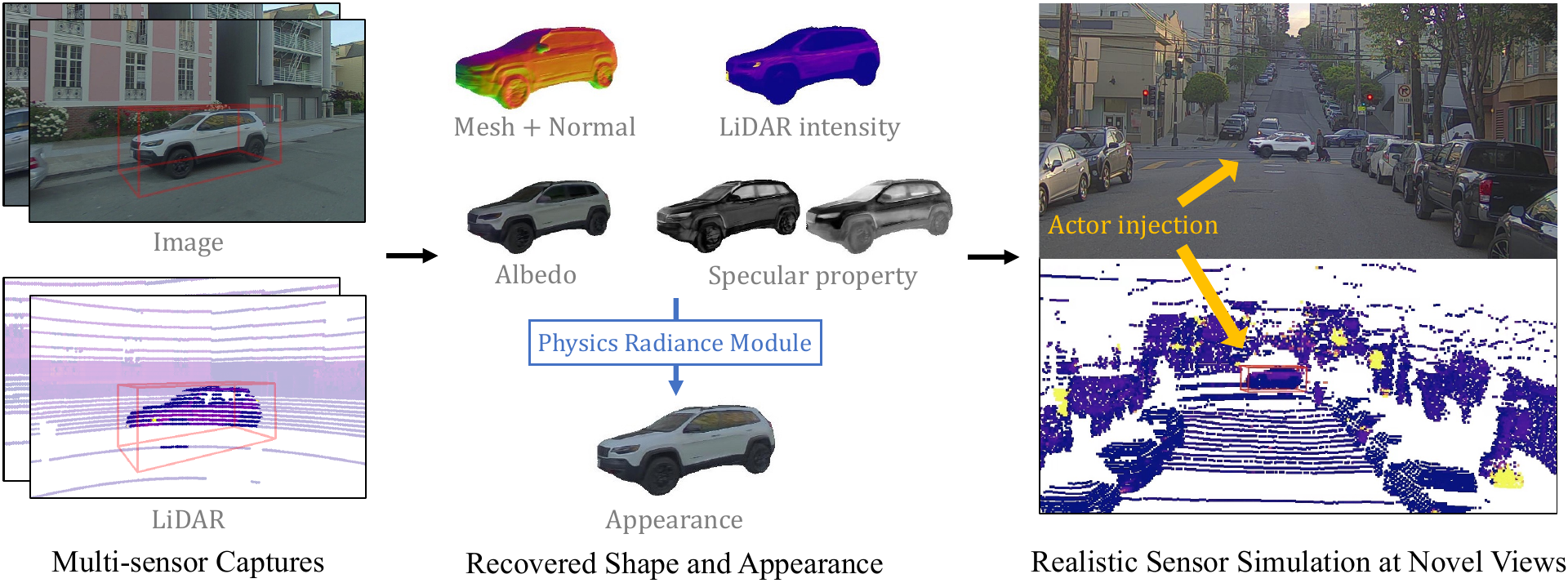}
    \end{center}
    \vspace{-5.5mm}
	\captionof{figure}{Given a camera video and LiDAR sweeps as input, our model reconstructs accurate geometry and surface properties, which can be used to synthesize realistic appearance under novel viewpoints using our physics-based radiance module, enabling realistic sensor simulation for self-driving.}
    \label{fig:teaser}
	\vspace{3.0mm}
}]

\begin{abstract}
Reconstructing objects from real world data and rendering them at novel views is
critical to bringing realism, diversity and scale to simulation for robotics
training and testing.
In this work, we present \name, a novel approach that estimates accurate geometry and realistic appearance from sparse in-the-wild data captured at distance and at limited viewpoints.
Towards this goal, we represent the object surface as a neural signed distance function and leverage both LiDAR and camera sensor data to reconstruct smooth and accurate geometry and normals.
We model the object appearance with a robust physics-inspired reflectance representation effective for in-the-wild data.
Our experiments show that \name has strong view synthesis performance on challenging scenarios with sparse training views.
Furthermore, we showcase composing \name assets into a virtual world and generating realistic multi-sensor data for evaluating self-driving perception models.
Please refer to our project website \url{https://waabi.ai/neusim/} for more results.
\end{abstract}

\vspace{1pt}
\section{Introduction}
\cutsectiondown

Simulation is key for testing at scale self-driving systems.
To allow for end-to-end closed loop testing, the simulator should produce in real-time realistic sensor data, which are rendered views of a 3D virtual world.
Most simulators employ CAD models~\cite{dosovitskiy2017carla,shah2018airsim}, which have unrealistic appearance, require costly manual construction and cannot scale to represent the diversity and complexity of the real world.
To address this issue, we focus on automatically reconstructing high-quality 3D assets cost-efficiently from sparse, in-the-wild, multi-sensor data captured by a moving platform along constrained trajectories.
The reconstructed assets should have accurate shape and appearance, and should render efficiently.

\begin{figure}[t]
	\begin{center}
		\includegraphics[width=0.44\textwidth]{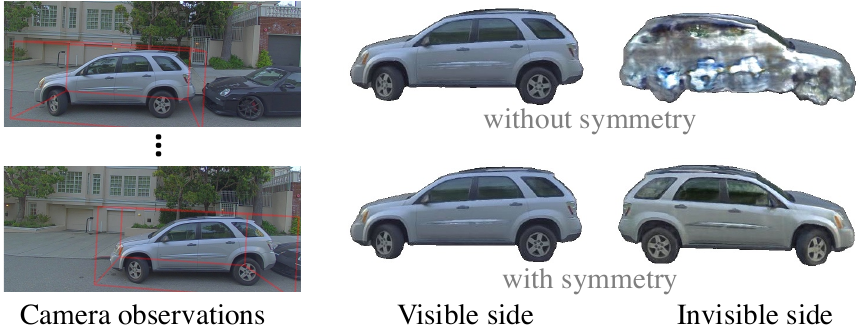}
	\end{center}
	\cuthalfcaptionup
	\caption{The structural symmetry priors helps to reconstruct shape and appearance on unseen regions for common traffic objects (\eg. cars, trucks).}
	\cuthalfcaptiondown
	\label{fig:symmetry_learning}
\end{figure}

\begin{figure}[t]
	\begin{center}
		\includegraphics[width=0.44\textwidth]{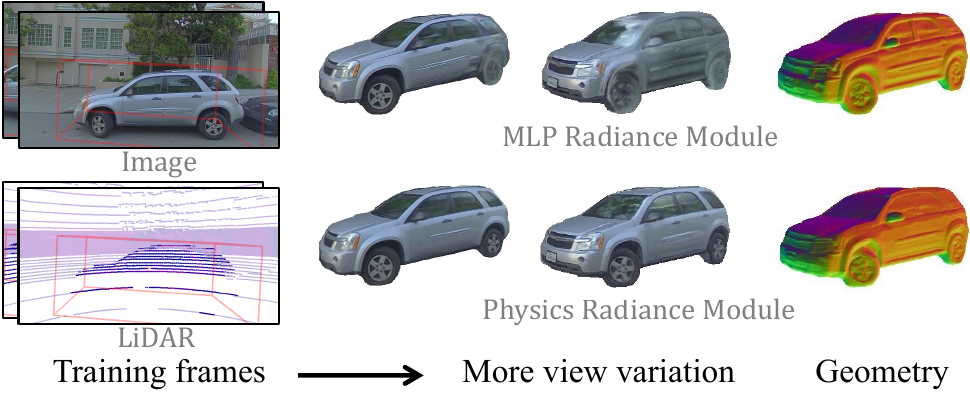}
	\end{center}
	\cuthalfcaptionup
	\caption{
	Prior MLP radiance model~\cite{mildenhall2020nerf,Oechsle2021ICCV,wang2021neus} cannot generalize well to large viewpoint change though the underlying geometry is correct.
	\name's Physics Radiance Module models the appearance robustly and realistically.}
	\cuthalfcaptiondown
	\vspace{-5pt}
	\label{fig:view_variation}
\end{figure}

A recent promising approach to asset shape reconstruction and novel view synthesis is Neural Radiance Fields (NeRF)~\cite{mildenhall2020nerf}, which represents the scene as a continuous density and radiance field parameterized by neural networks, and leverages volume rendering~\cite{kajiya1984ray} to render the scene. 
NeRF produces renderings that match the observed images;  however, the resulting asset representation is not suitable for scalable virtual world creation from in-the-wild data due to shape radiance ambiguity~\cite{zhang2020nerf++}, noisy geometry~\cite{wang2021neus,Oechsle2021ICCV}, incomplete reconstruction (see Fig.~\ref{fig:symmetry_learning}) and significant artifacts and distortion at extrapolated viewpoints (see Fig.~\ref{fig:view_variation}).
Several works have aimed to improve NeRF~\cite{zhang2020nerf++,jain2021putting,wang2021neus,Oechsle2021ICCV} and adapt it to outdoor scenes~\cite{zhang2020nerf++,barron2021mipnerf360}, but they focus on either synthetic data or small scenes densely captured in controlled environments.
In contrast, we focus on reconstruction from data captured by moving platforms (\eg, self-driving vehicle) in outdoor environments, which are more challenging due to the sparsity and limited range of sensor viewpoints, varying resolution and distance from the sensor, and sensor noise. 

In this paper, we propose \name, a novel neural volume rendering approach that takes sparse, in-the-wild image and LiDAR data and learns an asset's shape and surface properties for robust and realistic multi-sensor simulation.
\name is composed of a neural geometry representation that generates precise surfaces and models appearance via a physics-based radiance model, which accurately learns texture and reflectance.
\name also incorporates structural symmetry priors for common traffic actors (primarily cars, trucks) to learn the surface properties of the unobserved side (see Fig.~\ref{fig:symmetry_learning}), enabling robust novel view synthesis and seamless simulation of new scenarios.
\name not only renders and learns images, but also predicts LiDAR depths and intensities, resulting in better geometry than assets generated from images alone, and enabling consistent multi-sensor simulation, which is key for self-driving testing, as modern perception systems exploit multiple sensors for robustness.
Our factorized geometry and radiance model easily bakes into an explicit mesh allowing for easy modification, (1000x) faster rendering, and integration into existing simulators.

Our experiments show that \name reconstructs high-quality assets from in-the-wild data that render efficiently in existing graphics engines.
Finally, we leverage \name assets to generate sensor data for testing perception systems.

\cutsectionup
\section{Related Work}
\cutsectiondown

\subsubsection{Assets for Sensor Simulation}
Self-driving simulators such as CARLA~\cite{dosovitskiy2017carla} and AirSim~\cite{shah2018airsim} leverage artist-created 3D CAD models to build virtual worlds.
While these assets have clean geometry and are easily modifiable, they are expensive to create, have limited diversity and scale, and lack realism~\cite{hoffman2018cycada}.
Several works leverage in-the-wild data from driving scenes to build assets at scale.
LiDARSim and SurfelGAN~\cite{manivasagam2020lidarsim,yang2020surfelgan} aggregate LiDAR across frames to build diverse textured surfel representations, but have noisy geometry and appearance.
Other works perform shape completion and texture estimation using a learned asset representation.
Several leverage synthetic CAD data~\cite{engelmann2017samp,lu2020permo,yang2020recovering,duggal2021secrets,zhao2020vehicle} as a prior, and then generate the final asset either through feed-forward prediction or optimization on real data.
These approaches are limited by the set of CAD models used and are often not photorealistic.
GeoSim~\cite{chen2021geosim} deforms a template mesh and leverages image warping for photorealistic video simulation, but the assets do not model reflectance and have overly smooth geometry due to its limited mesh representation.
DriveGAN~\cite{kim2021drivegan} represents assets as disentangled latent codes and generates video from control inputs, allowing for fully differentiable simulation, but is limited in its realism and is not temporally consistent.
In contrast, we build realistic assets that have high-fidelity geometry, can be inserted into new and diverse scenarios at scale, and are temporally and multi-sensor consistent.

\begin{figure*}[t]
    \begin{center}
        \includegraphics[width=0.95\textwidth]{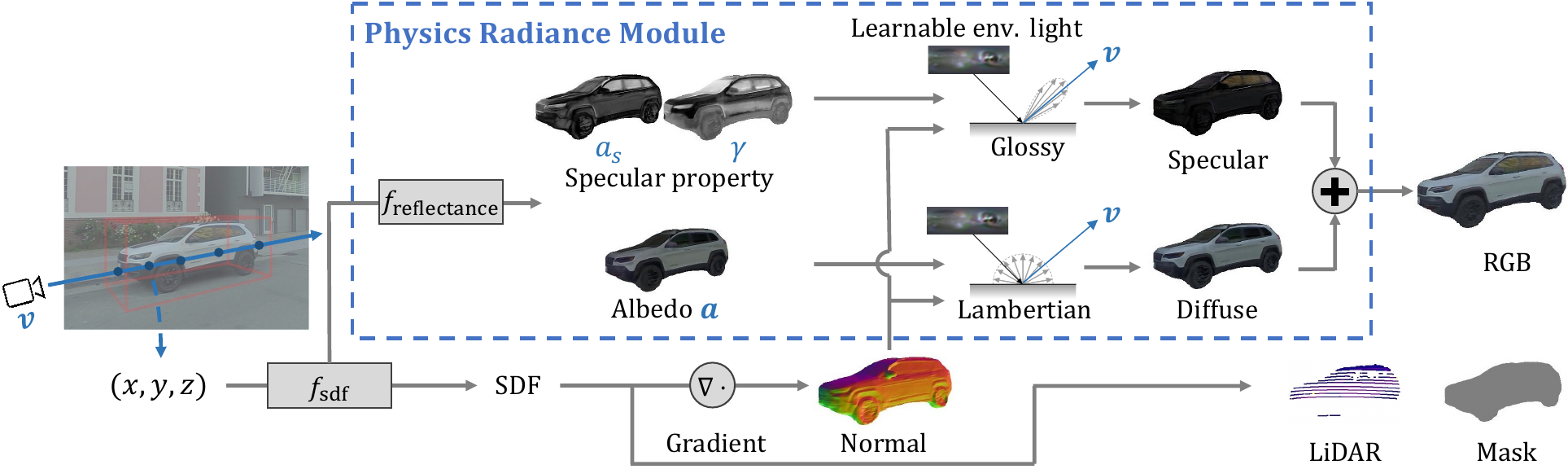}
    \end{center}
    \cuthalfcaptionup
	\caption{
	Given a continuous 3D location, \name~outputs the signed distance value of the point to object surface, the albedo, and the specular property. 
	The signed distance value is used to derive the surface normal, which is then used to shade the diffuse and specular components to obtain the final RGB color. 
	We also render the LiDAR depth and intensity, as well as object mask from the learned representation.
	}
    \cuthalfcaptiondown
    \vspace{-5pt}
    \label{fig:overview}
\end{figure*}

\subsubsection{Neural Volume Rendering}
Neural implicit representation~\cite{park2019deepsdf,chen2019learning,yang2021s3,mescheder2019occupancy,atzmon2020sal,davies2020effectiveness,sitzmann2021awesome,sitzmann2019scene,sitzmann2020implicit} and volumetric rendering approaches~\cite{niemeyer2020differentiable,kajiya1984ray}, such as Neural Radiance Fields (NeRF)~\cite{mildenhall2020nerf} have achieved high quality image rendering given dense and uniform training views, but can have noisy geometry and artifacts, especially when trained on sparse and limited viewpoints.
To learn a better geometry, recent works have assumed solid closed surface objects and combined SDFs with radiance fields~\cite{yariv2021volume,azinovic2021neural,wang2021neus,Oechsle2021ICCV,insafutdinov2022snes} and leverage 3D information such as RGB-D measurements~\cite{deng2021depth,azinovic2021neural}, resulting in improved geometry.
We also leverage an SDF representation, and incorporate LiDAR data as supervision.
Our work combines these enhancements with a learned reflectance model to learn more robust appearance (see Fig.~\ref{fig:view_variation}) for in-the-wild data.

Several works learn a reflectance model and estimate scene lighting~\cite{nerfactor,srinivasan2021nerv,bi2020neural,boss2021nerd,zhang2021physg}, but they focus on improving controllability of NeRF and learn on synthetic data or controlled environments.
We leverage a simple yet effective Phong illumination model to model reflectance~\cite{phong1975illumination,arvo1995applications}, and supervise with multi-sensor data (RGB and LiDAR intensity), to better render objects at novel views from sparse in-the-wild data. 
We also incorporate structural priors such as symmetry over the asset shape and surface properties to complete the missing information due to limited sensor observations.

Other works have focused on the composability of neural radiance fields~\cite{ost2021neural,niemeyer2021giraffe,liu2021editing,yang2021learning,schwarz2020graf,guo2020object}, but do not model the reflectance and do not generate 3D geometry for multi-sensor simulation.
In contrast, our approach models reflectance and can be baked into an explicit textured mesh for fast rendering.

\subsubsection{Inverse Graphics}
Several recent works generate assets from in-the-wild data with decomposed shape, material, and illumination via differentiable mesh rendering~\cite{chen2021dib,zhang2021physg,zhang2021ners}.
These approaches build on intrinsic decomposition, the task of decomposing the image into albedo, illumination, etc. based on image formation~\cite{barrow1978recovering,barron2013intrinsic,ma2018single}.
Our work is most similar to NeRS~\cite{zhang2021ners}, which estimates shape through a deformable implicit surface representation and appearance with a learned Phong reflectance model.
However, our use of volume rendering and an SDF representation results in more accurate assets, as shown in our experiments.

\cutsectionup
\section{Method}
\cutsectiondown

Our goal is to automatically build rigid object assets from in-the-wild data for sensor simulation.
Given camera images and LiDAR point clouds captured by a moving platform, we want to learn the object's shape and appearance.
Towards this goal, we propose \name, a novel approach that is composed of a structured neural surface representation and a physics-based reflectance model.
This decomposed representation enables generalization to new views from sparse in-the-wild viewpoints.
Fig.~\ref{fig:overview} provides an overview of \name.
First, we briefly review  NeRF.
Next, we introduce Neural Surface Modeling for accurate surface and normal estimation.
We then introduce our Physics-based Radiance Module for robust and realistic appearance modelling.
Finally we present how our model performs rendering and learns from data.

\subsection{Review of NeRF Representations}
\label{sec:nerf}
NeRF~\cite{mildenhall2020nerf} represents the scene with a Multi-Layer Perceptron (MLP) that maps a point location $\bx \in \mathbb R^3$ and viewing direction $\bd \in \mathbb R^3$ to a volume density $\sigma(\bx) \in \mathbb R$ and RGB color radiance $\bc (\bx, \bd) \in \mathbb R^3$. 
The color is conditioned on the viewing direction to model view-dependent effects.
Given a posed camera, NeRF performs volume rendering by evaluating the MLP at points along a camera ray to compute pixel color.
We define each pixel's ray as $\{ \mathbf r(t) = \bo + t\bd \  | \  t\!>\!0 \}$, where $\bo$ is camera center and $\bd$ is the viewing direction originating from the camera center. 
Given $N$ sample points $\{ \bx_i \}_{i=1}^N$ along the camera ray, the pixel color is
\vspace{-2pt}
\begin{align}
	\label{eqn:camera_rendering}
	\mathbf C(\mathbf r) &= \sum\limits_{i=1}^N \alpha(\mathbf x_i) T(\mathbf x_i) \mathbf c(\mathbf x_i, \mathbf d)\\
	\alpha(\mathbf x_i) &= 1-\exp\left(-\sigma(\mathbf x_i)\delta_i\right), \quad T(\mathbf x_i) = \prod_{j=1}^{i-1} (1-\alpha(\mathbf x_j)) \nonumber
\end{align}
\vspace{-2pt}
where $\alpha(\mathbf x_i)$ is the alpha compositing computed using the density prediction $\sigma(\mathbf x)$ and the distance between adjacent samples $\delta_i = \left\lVert \mathbf x_{i+1} - \mathbf x_i \right\rVert_2$, and $T(\mathbf x_i)$ is the accumulated transmittance along the ray.

\subsection{Neural Surface Modeling}
\label{sec:geo}
NeRF-based approaches \cite{mildenhall2020nerf,zhang2020nerf++,barron2021mip} generate photorealistic renderings, but are unable to reconstruct accurate geometry, especially when trained on sparse in-the-wild views. 
Rather than model the space as a heterogenous soft volume as in NeRF~\cite{mildenhall2020nerf}, we take inspiration from~\cite{park2019deepsdf,yariv2021volume,azinovic2021neural,Oechsle2021ICCV} and assume the object of interest has a topologically closed surface that we represent as the zero-level set of a signed distance function (SDF), parameterized by an MLP.
This representation is effective for most objects of interest in self-driving simulations (e.g., vehicles, construction elements).

We define an SDF MLP network $f_{\text{SDF}}: \mathbf x \rightarrow s$ mapping a point in 3D space $\mathbf x \in \mathbb R^3$ to its signed distance $s(\mathbf x) \in \mathbb R$ from the object surface.
The surface $\mathcal S$ of the object can then be defined as the zero-level set of the SDF function:
\begin{align}
	\mathcal S = \{ \mathbf x \in \mathbb R^3 \; | \; s(\mathbf x) = 0 \}.
\end{align}
And the normal is derived as the SDF's gradient, $\mathbf n = \nabla s(\mathbf x)$.

To perform volume rendering (Eq.~\eqref{eqn:camera_rendering}), we need to convert the signed distance to an alpha value $\alpha(\mathbf x) \in [0, 1]$, where $\alpha(\mathbf x) = 1$ if  $\mathbf x$ is inside the object, and $\alpha(\mathbf x) = 0$ otherwise. 
Towards this goal, we use a sigmoid-like function:
\begin{align}
	\label{eqn:sdf_to_alpha}
	\alpha(\mathbf x) =  \frac{1}{1+\exp(\beta \cdot s(\mathbf x))},
\end{align}
where $\alpha(\mathbf x)$ transitions from $0$ to $1$ when $s(\mathbf x)$ transitions from positive (outside) to negative (inside).
To reconstruct solid objects, the converting function should model a step function at the object's surface, \ie, $\beta \to \infty$.
To prevent vanishing gradients, we make $\beta$ a learnable parameter.

\cutsubsectionup
\subsection{Robust and Realistic Appearance Modelling}
\label{sec:reflect}
\cutsubsectiondown
Given our surface geometry representation, we now discuss how we model object appearance. 
We represent the appearance of a 3D point $\mathbf x$ viewed from direction $\mathbf d$ as a radiance field $\mathbf c(\mathbf x, \mathbf d)$, a function of the material properties $f_r$, the radiance from the environment $L(\mathbf x, \omega)$ (with $\omega$ the incoming radiance direction), and the surface normals $\mathbf n(\mathbf x)$.
$\mathbf c(\mathbf x, \mathbf d)$ is estimated via rendering equation~\cite{immel1986radiosity,kajiya1986rendering}:
\begin{align}
	\label{eqn:reflectance_equation}
	\mathbf c(\mathbf x, \mathbf d) &=
	\int_\Omega f_r(\mathbf x, \omega, \mathbf d) L(\mathbf x, \omega)(\omega \cdot \mathbf n) \mathrm d \omega
\end{align}
where $f_r(\mathbf x, \omega, \mathbf d)$ is the bidirectional reflectance distribution function (BRDF), which defines the proportion of light reflected from the incoming light direction $\omega$ to the viewing direction $\bd$ at position $\bx$.
The integral is computed over the hemisphere $\Omega$ centered at location $\mathbf x$ and oriented towards the normal $\mathbf n$. 
Since in outdoor settings the light mainly comes from the sun, we model the scene lighting as a set of infinitely far-away directional lights, and ignore attenuation effects to the incoming light intensity.
Thus the light function is independent of the point location, \ie, $L(\mathbf x, \omega) = L(\omega)$.

NeRF models the radiance equation using an MLP that directly maps a point $\bx$ and view direction $\bd$ to the emitted radiance $\bc(\bx, \bd)$.
This does not generalize well to larger viewpoint variation (see Fig.~\ref{fig:view_variation}), especially when trained on a sparse set of images.
The MLP radiance lacks knowledge of the underlying light transport physics
and overfits.
To address this, we model the radiance function using a simple yet effective Phong illumination model~\cite{phong1975illumination,arvo1995applications} with spatial varying albedo and shininess.
Our framework also supports other PBR models (e.g., microfacet~\cite{walter2007microfacet}).
The Phong model consists of diffuse and specular components:
\begin{align}
	\label{eqn:phong_integral}
	\mathbf c_\text{phy}(\mathbf x, \mathbf d) = \int_\Omega
	\Big(
	\underbrace{\mathbf a (\mathbf x) (\omega \cdot \mathbf n)}_{\text{Diffuse term}} +
	\underbrace{f_s(\mathbf x, \mathbf d)}_{\text{Specular term}}
	\Big)
	\underbrace{L(\omega)}_{\text{Light}}
	\mathrm d \omega
\end{align}
The diffuse component depends on the albedo $\mathbf a (\mathbf x)\in \mathbb R^3$.
The specular component $f_s(\mathbf x, \mathbf d)$ is view-dependant:
\begin{align}
	f_s(\mathbf x, \mathbf d) = a_s (\mathbf x) \frac{\gamma(\mathbf x) +1}{2\pi} (\mathbf r_{\omega} \cdot \mathbf d)^{\gamma(\mathbf x)}
\end{align}
where $a_s (\mathbf x)\in \mathbb R$ is the specular albedo measuring the specular highlight intensity, $\mathbf r_{\omega} = 2(\omega\cdot \mathbf n)\mathbf n - \omega$ is the reflected light direction, and $\gamma(\mathbf x)$ is the shininess.

\begin{figure}[t]
	\begin{center}
		\includegraphics[width=0.45\textwidth]{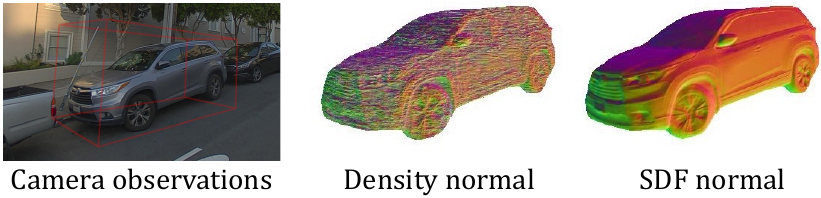}
	\end{center}
	\cuthalfcaptionup
	\caption{Rendering of normal derived from density and SDF representation.}
	\cuthalfcaptiondown
	\cuthalfcaptiondown
	\label{fig:accurate_normal}
\end{figure}

We predict the diffuse albedo $\mathbf a (\mathbf x)$, specular albedo $a_s (\mathbf x)$ and shininess $\gamma(\mathbf x)$ using reflectance MLPs $f_{\text{reflectance}}$.
We use a learnable 2D environment map $\mathcal E \in \mathbb R^{A \times E \times 3}$ to parameterize the environment light $L(\omega)$, where $A$ and $E$ are the azimuth and elevation resolution respectively.
$\mathcal E$ stores the light intensity for each discrete incoming light direction $\omega$.
We numerically estimate integral in Eq.~\eqref{eqn:phong_integral} as the sum of radiance values from discrete incoming lights $\mathcal E_{\omega} = L(\omega)$:
\begin{align}
	\label{eqn:phong_radiance}
	\mathbf c_\text{phy} &\approx \mathbf a \sum\limits_{\omega} (\omega \cdot \mathbf n) \mathcal E_{\omega}
	+  \frac{a_s(\gamma +1)}{2\pi} \sum\limits_{\omega}  (\mathbf r_{\omega} \cdot \mathbf d)^{\gamma} \mathcal E_{\omega}
\end{align}
$\mathbf c_\text{phy}(\mathbf x, \mathbf d)$ relies on accurate normal estimation $\mathbf n$.
Prior NeRF-based methods estimates the normal~\cite{nerfactor,srinivasan2021nerv,boss2021nerd} by taking the normalized gradient of volume density \wrt ~input 3D location, but they are noisy (see Fig.~\ref{fig:accurate_normal}).
Other methods use a MLP to predict the normal vector and enforce smoothness via regularization~\cite{nerfactor,bi2020neural}.
These predicted normals tend to be smoother, but they are decoupled from the actual shape and do not reflect the exact geometry.
In contrast, our neural SDF representation ensures smooth and accurate normal estimated as the SDF's gradient, $\mathbf n = \nabla s(\mathbf x)$.

\cutsubsectionup
\subsection{Rendering with \name}
\label{sec:rendering}
\cutsubsectiondown

\subsubsection{Scene Representation}
As we focus on object reconstruction, we assume that the object of interest in the scene is bounded by a cuboid and we only render in the frustum generated between the viewing plane and the projected cuboid~\cite{ost2021neural}. 
We assume the rendered ray $r(t)$ intersects the cuboid at $t_\text{near}$ and $t_\text{far}$.
We divide the traversed space into foreground ($\{t_\text{near}<t<t_\text{far} \}$) and background.
To decouple the object sensor observations from the background's, we take inspiration from NeRF++~\cite{zhang2020nerf++} and model them with separate networks. 
We use our proposed model to represent the foreground and use NeRF with an inverted sphere parameterization~\cite{zhang2020nerf++} to represent the background.
For rendering only the asset, the background network is discarded.

\subsubsection{Rendering Camera}
To render the RGB observations, we first draw stratified samples on the ray and query the SDF MLPs to compute $\alpha(\mathbf x)$ as in Eq.~\eqref{eqn:sdf_to_alpha}. 
For each sampled point, we query the reflectance MLPs to compute the albedo $\mathbf a(\mathbf x), a_s(\mathbf x)$ and shininess $\gamma(\mathbf x)$. 
We then compute the foreground radiance $\mathbf c(\mathbf x, \mathbf d)$ for each sampled point $\mathbf x$ with view direction $\mathbf d$, using Eq.~\eqref{eqn:phong_radiance} and the learned environment map $\mathcal E$.
To compute the background scene radiance, we sample the ray's intersections with Multiple-Sphere Images (MSI) surrounding the object of interest. 
We generate the radii for MSI by linearly interpolating inverse depths.
The rendered RGB is computed by alpha compositing using Eq.~\eqref{eqn:camera_rendering}.

\subsubsection{Rendering LiDAR Depth and Intensity}
Given a query LiDAR ray, we sample points along the ray similar to camera rendering, and then render the LiDAR depth and intensity as:
\begin{equation}
\label{eqn:depth_rendering}
\resizebox{.43\textwidth}{!}{$
\begin{split}
	D(\mathbf r) = \sum\limits_{i=1}^N \alpha(\mathbf x_i) T(\mathbf x_i) d_i,\quad I(\mathbf r) = \sum\limits_{i=1}^N \alpha(\mathbf x_i) T(\mathbf x_i) i(\mathbf x).
\end{split}$}	
\end{equation}
where $d_i = \left\lVert \mathbf x_i - \mathbf o \right\rVert_2$ is the depth value of sample point $\mathbf x_i$ on the ray originating from $\mathbf o$.
We add another branch on the reflectance MLP to predict the intensity value $i(\mathbf x) \in \mathbb R^+$.

\subsubsection{Rendering Object Mask}
We render the object mask to help provide additional signal on object boundaries.
We estimate the rendered foreground probability as the aggregated weights in the foreground intervals:
\begin{align}
	m(\mathbf r) = \sum_{t_\text{near}<t<t_\text{far}} \alpha(\mathbf o + t \mathbf d) T(\mathbf o + t \mathbf d).
\end{align}

\cutsubsectionup
\subsection{Learning \name}
\label{sec:learning}
\cutsubsectiondown
To learn the model, we minimize the difference between the sensor observations and our rendered outputs.
We leverage RGB color ($\cL_\text{color}$), LiDAR point clouds ($\cL_\text{lidar}$) and object foreground masks ($\cL_\text{mask}$). 
We also add an Eikonal term ($\cL_\text{Eik}$) to regularize the predicted SDF, and a symmetry term ($\cL_\text{sym}$) for vehicle objects.
The full training loss is:
\begin{equation}
\label{eqn:total_loss}
\resizebox{.43\textwidth}{!}{$
\begin{split}
	\mathcal L = \mathcal L_\text{color} + \lambda_\text{lidar} \mathcal L_\text{lidar} + \lambda_\text{mask} \mathcal L_\text{mask} + \lambda_\text{Eik} \mathcal L_\text{Eik} + \lambda_\text{sym} \mathcal L_\text{sym}
\end{split}$}	
\end{equation}
For each asset, we train the shape and reflectance networks and environment lighting map jointly via gradient descent over ray batches of size $N$ randomly sampled from the sensor data.
We now review each loss term.

\subsubsection{RGB supervision}
Similar to NeRF, we want to ensure that the rendered pixels match the observed ones.
The camera image loss $\mathcal L_\text{color}$ is defined as:
\vspace{-2pt}
\begin{align}
	\mathcal L_\text{color} = \frac{1}{N} \sum\limits_{i=1}^N \left\lVert \mathbf C(\mathbf r_i) - \mathbf{\hat{C}}_i \right\rVert_1
\end{align}
\vspace{-2pt}
where $\mathbf C(\mathbf r_i)$ is rendered color and $\mathbf{\hat{C}}_i$ is observed color.

\subsubsection{LiDAR supervision}
We leverage LiDAR depth measurements to supervise the SDF field for more accurate geometry, as well as intensity to learn better shape and surface properties.
The LiDAR loss $\mathcal L_\text{lidar}$ is defined as:
\begin{align}
	\mathcal L_\text{lidar} \!=\! \frac{1}{N} \sum\limits_{i=1}^N \! \left( \left\lVert D(\mathbf r_i) \!-\! \hat{D}_i \right\rVert_2
	+\lambda_\text{int} \left\lVert I(\mathbf r_i) \!-\! \hat{I}_i \right\rVert_2 \right)
\end{align}
where $D(\mathbf r_i)$ is the rendered LiDAR depth and $I(\mathbf r_i)$ is the rendered LiDAR intensity. $\hat{D}_i$ and $\hat{I}_i$ are the LiDAR depth and intensity observations, respectively.
We also penalize large weight predictions that are far from the observations:
\vspace{-2pt}
\begin{align}
\mathcal L_\text{lidar}= \mathcal L_\text{lidar} + \frac{1}{N} \sum\limits_{i=1}^N \left( \sum_{\left\lVert d_{ij}-\hat{D}_i \right\rVert_2 > \epsilon} w_{ij} \right)
\end{align}
\vspace{-2pt}

\subsubsection{Object mask supervision}
Foreground masks identify the object in the image.
The mask loss $\mathcal L_\text{mask}$ is defined as:
\begin{align}
	\mathcal L_\text{mask} = \frac{1}{N} \sum\limits_{i=1}^N \left\lVert m(\mathbf r_i) - \hat{m}_i \right\rVert_2,
\end{align}
where $m(\mathbf r_i)$ is the rendered foreground probability and $\hat{m}_i$ is the estimated mask from an off-the-shelf algorithm~\cite{kirillov2020pointrend}.

\subsubsection{Eikonal Regularizer}
This term encourages the SDF to satisfy the Eikonal equation and generate unit normals~\cite{sitzmann2020implicit}:
\begin{align}
	\mathcal L_\text{Eik} &= \frac{1}{M} \sum\limits_{i=1}^M \left( \left\lVert \nabla s(\mathbf x_{i}) \right\rVert_2 - 1 \right)^2
\end{align}
where $\{ s(\mathbf x_{i}) \}_{i=1}^M$ is the predicted signed distance in a batch.

\subsubsection{Structural Symmetry Prior}
To reconstruct unseen regions, we incorporate symmetry priors for common traffic objects (\eg, cars, trucks).
Although the radiance is not symmetric due to diffuse and specular shading with lighting, the surface geometry $s(\mathbf x)$ and material properties ($\mathbf a(\mathbf x), a_s(\mathbf x), \mathbf\gamma(\mathbf x)$) are approximately symmetric.
We denote the transform from world coordinate to the canonical object coordinate (Front-Left-Up) as 
$\mathbf T =
\big(\begin{smallmatrix}
    \mathbf R & \mathbf t \\
    \mathbf{0} & 1
\end{smallmatrix}\big)  
\in SE(3)$. 
For each query point $\mathbf x$ with normal $\mathbf n$ in world coordinate, the symmetrized point $\mathbf x'$ and normal $\mathbf n'$ are:
\begin{equation}
\resizebox{.43\textwidth}{!}{$
\begin{split}
\mathbf x' = \mathbf T^{-1} 
\begin{pmatrix}
1 & 0 & 0 & 0\\
0 & -1 & 0 & 0\\
0 & 0 & 1 & 0\\
0 & 0 & 0 & 1
\end{pmatrix} 
\mathbf T \mathbf x,\quad
\mathbf n' = \mathbf R^{-1} 
\begin{pmatrix}
1 & 0 & 0 \\
0 & -1 & 0 \\
0 & 0 & 1 \\
\end{pmatrix} 
\mathbf R \mathbf n
\end{split}$}	
\end{equation}
We jointly optimize $\mathbf T$ during training, the symmetry loss is:
\begin{equation}
\resizebox{.43\textwidth}{!}{$
\begin{split}
	\mathcal L_\text{sym} 
	& = \frac{1}{M} \sum\limits_{i=1}^N \left\lVert s(\mathbf x_i) - s(\mathbf x'_i)\right\rVert_2 + \left\lVert \mathbf n' (\mathbf x_i) -  \mathbf n(\mathbf x'_i)\right\rVert_2 \\
	& + \left\lVert \mathbf a(\mathbf x_i) -  \mathbf a(\mathbf x'_i)\right\rVert_2 + \left\lVert a_s(\mathbf x_i) -  a_s(\mathbf x'_i)\right\rVert_2 + \left\lVert \mathbf \gamma(\mathbf x_i) - \gamma(\mathbf x'_i)\right\rVert_2
\end{split}$}	
\end{equation}

\begin{table}[t]
    \begin{center}
    \begin{tabular}{lcccc}
	\toprule[0.1em]
    Method & MSE$\downarrow$ & PSNR$\uparrow$ & SSIM$\uparrow$ & LPIPS$\downarrow$ \\
    \midrule
    \midrule
    SI-ViewWarp~\cite{tulsiani2018layer} & 0.0233 & 17.51 & 0.514 & 0.371 \\
    SAMP~\cite{engelmann2017samp} & 0.0144 & 19.52 & 0.628 & 0.283 \\
    \midrule
    NeRS~\cite{zhang2021ners} & 0.0176 & 18.49 & 0.562 & 0.265 \\
    NVDiffRec~\cite{munkberg2022extracting} & 0.0114 & 20.46 & 0.593 & 0.396 \\
    \midrule
    NeRF++~\cite{zhang2020nerf++} & 0.0138 & 20.86 & 0.611 & 0.300 \\
    NeuS~\cite{wang2021neus} & 0.0115 & 21.37 & 0.640 & 0.247 \\
    \rowcolor{grey} Ours & \textbf{0.0081} & \textbf{22.44} & \textbf{0.692} & \textbf{0.202} \\
	\bottomrule[0.1em]
    \end{tabular}
    \end{center}
    \cuthalftablecaptionup
    \caption{Quantitative comparisons on PandaVehicle for NVS task.}
    \cuthalftablecaptiondown
    \vspace{-14pt}
    \label{tab:sota_comparison}
\end{table}

\begin{table}[t]
    \begin{center}
    \begin{tabular}{lcccc}
	\toprule[0.1em]
    Supervision & MSE$\downarrow$ & PSNR$\uparrow$ & SSIM$\uparrow$ & LPIPS$\downarrow$ \\
    \midrule
    img & 0.0131 & 21.07 & 0.646 & 0.269 \\
    img + lidar & 0.0089 & 22.07 & 0.677 & 0.211 \\
    img + lidar + mask & 0.0091 & 22.13 & 0.679 & \textbf{0.198} \\
    img + lidar + mask + sym & \textbf{0.0081} & \textbf{22.44} & \textbf{0.692} & 0.202 \\
	\bottomrule[0.1em]
    \end{tabular}
    \end{center}
    \cuthalftablecaptionup
    \caption{Ablation on learning supervision for \name.}
    \cuthalftablecaptiondown
    \label{tab:ablate_supervision}
\end{table}

\begin{table}[t]
    \begin{center}
    \begin{tabular}{lcccc}
	\toprule[0.1em]
    Radiance Model & MSE$\downarrow$ & PSNR$\uparrow$ & SSIM$\uparrow$ & LPIPS$\downarrow$ \\
    \midrule
    MLP Radiance & 0.0118 & 21.31 & 0.636 & 0.232 \\
    \rowcolor{grey} Physics-based Radiance & \textbf{0.0091} & \textbf{22.13} & \textbf{0.679} & \textbf{0.198} \\
	\bottomrule[0.1em]
    \end{tabular}
    \end{center}
    \cuthalftablecaptionup
    \caption{Ablation on radiance model for appearance prediction.}
    \cuthalftablecaptiondown
    \vspace{-12pt}
    \label{tab:ablate_reflectance}
\end{table}

\cutsectionup
\section{Experimental Evaluation}
\cutsectiondown
In this section we demonstrate our model performance on in-the-wild data.
We first introduce our experimental settings.
We then compare our model against state-of-the-art methods, and also ablate our design choices.
Finally, we apply \name for fast and realistic sensor simulation for self-driving.

\begin{figure*}[t]
    \begin{center}
        \includegraphics[width=1.0\textwidth]{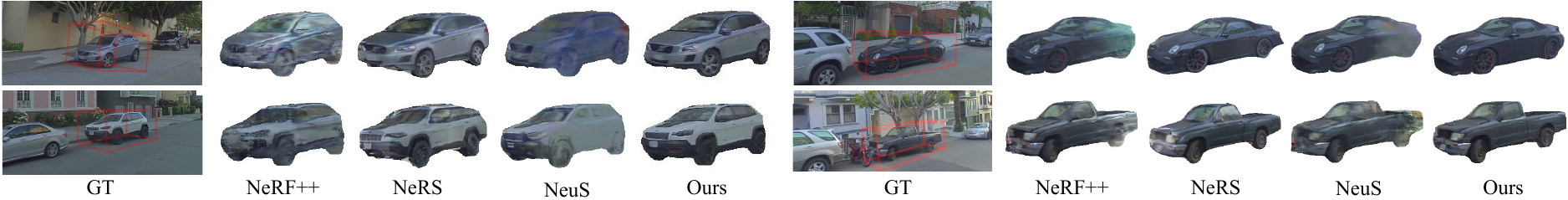}
    \end{center}
    \cuthalfcaptionup
    \caption{Qualitative comparisons on PandaVehicle for novel view synthesis. For each vehicle, we show results at novel views with large view variation.}
    \label{fig:sota_comparison}
\end{figure*}

\begin{figure*}[t]
    \begin{center}
        \includegraphics[width=1.0\textwidth]{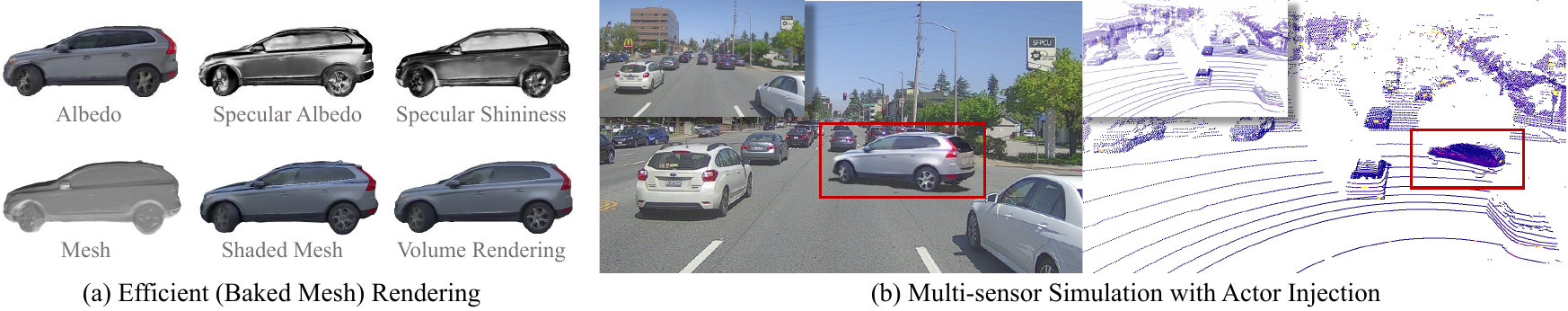}
    \end{center}
    \cuthalfcaptionup
    \caption{\name reconstructs disentangled shape and appearance, enabling (a) efficient mesh rendering by baking diffuse and specular properties into an explicit mesh, which is 1000x faster than volume rendering, (b) and multi-sensor simulation for self-driving via actor insertion.
    }
    \label{fig:applications}
\end{figure*}

\cutsubsectionup
\subsection{Experimental Setting}
\cutsubsectiondown
We focus on recovering the shape and appearance of vehicles, the most common actor in self-driving scenes, and evaluate on the task of novel view synthesis.
We curated 10 vehicles with diverse shape and appearance under complex illumination from the PandaSet~\cite{xiao2021pandaset} to derive the PandaVehicle dataset.
This dataset has calibrated LiDARs and multiple cameras.
We use the left camera for training and the front-left camera for evaluation.
Each asset has on average $\sim$24 training views.
PandaVehicle is more challenging than existing novel view synthesis datasets due to limited range and number of viewpoints available for training, as well as the complex illumination.
We evaluate the NVS performance using Mean-Square Error (MSE), Peak Signal-to-Noise Ratio (PSNR), Structural Similarity Index Measure (SSIM), and Learned Perceptual Image Patch Similarity (LPIPS)~\cite{zhang2018perceptual}.
Since we focus on assets, we use the predicted segmentation mask~\cite{kirillov2020pointrend} to only evaluate the foreground pixels.

\cutsubsectionup
\subsection{Novel View Synthesis}
\cutsubsectiondown

\subsubsection{State-of-the-art (SoTA) Comparison}
We compare our model with state-of-the-art Neural Radiance Fields based approach NeRF++~\cite{zhang2020nerf++}, NeuS~\cite{wang2021neus} and inverse graphics model NeRS~\cite{zhang2021ners}, NVDiffRec~\cite{munkberg2022extracting}.
We choose these baselines as they model view-dependent appearance and work well in our outdoor setting.
We also compare against non-learning based LiDAR guided single view warping model~\cite{tulsiani2018layer} and CAD-based model SAMP~\cite{engelmann2017samp}.
As shown in Table~\ref{tab:sota_comparison}, our model achieves the best performance across all metrics.
A qualitative comparision is depicted by Fig.~\ref{fig:sota_comparison}.
Our model generalizes better to large viewpoint changes when compared to NeRF-based method~\cite{zhang2020nerf++, wang2021neus}, demonstrating the value of our physics-based reflectance module.
Our model also captures more fine-grained details than inverse graphics model~\cite{zhang2021ners} due to the expressive implicit representation.

\vspace{2pt}
\subsubsection{Ablation on Learning Supervision}
We study the effect of LiDAR, mask and symmetry supervision in Table~\ref{tab:ablate_supervision}.  
Incorporating LiDAR improves performance since the additional depth and intensity measurements help learn better geometry and reflectance.
Mask and symmetry supervision does not significantly improve metrics but it helps separate objects from the ground and complete unseen regions.

\vspace{2pt}
\subsubsection{Ablation on Radiance Model}
We study the effect of the radiance model in Table~\ref{tab:ablate_reflectance}.
For the MLP setting, we replace the reflectance model with a 4-layer viewpoint-dependant MLP predicting the color similar to~\cite{wang2021neus}.
Our reflectance model achieves better performance, demonstrating the benefits of a physics-based decomposed representation.

\cutsubsectionup
\subsection{Downstream Applications}
\cutsubsectiondown

\subsubsection{Efficient Rendering}
Our approach recovers geometry, surface texture and specular material that can be baked into an explicit mesh for efficient rendering.
We perform marching cubes to generate the mesh from the SDF representation~\cite{lorensen1987marching}, and evaluate on each vertex to compute the per-vertex albedo, specular albedo, and shininess mappings.
We render the explicit mesh with OpenGL using a customized shader implementing Eq.~\eqref{eqn:phong_radiance}.
In Fig.~\ref{fig:applications} (a), we show the reconstructed albedo and specular material, followed by a visual comparison between using mesh rendering and volume rendering.
Mesh rendering is 1000x faster than volume rendering ($\sim 77$ FPS vs. $\sim 0.03$ FPS on consumer GPU GTX 1080 Ti and Ubuntu OS) and still provides good visual quality. 
This enables efficient simulation.

\subsubsection{Downstream Evaluation on Camera Simulation}
We evaluate the object detection and instance segmentation algorithm on the simulated camera rendering at novel viewpoints.
Specifically, we insert and blend the actors into background image and replace the existing actors.
Then we compute the instance-level IoU of predicted bounding box and segmentation mask between real image and simulated image. 
This detection/segmentation agreement metric indicates how well we can use the sensor simulation to test existing perception systems. 
As shown in Tab.~\ref{tab:downstream_evaluation}, \name achieves highest agreement for both detection and instance segmentation.

\subsubsection{Realistic Sensor Simulation}
Using our reconstructed asset from \name, we can create consistent multi-sensor simulations for self-driving. 
For camera simulation, we render the asset to the target view and then apply a post-composition network~\cite{chen2021geosim} to seamlessly blend the actor to the background.
For LiDAR simulation, we use an approach similar to~\cite{fang2021lidar,wang2021advsim} and perform actor injection by raycasting the asset according to the LiDAR calibration and removing points in the real LiDAR sweep that are occluded by the added actor. 
Fig.~\ref{fig:applications} (b) shows that we can generate realistic camera and LiDAR simulations for the added actor, enabling diverse data generation and end to end autonomy testing.

\begin{table}[t]
    \begin{center}
	\begin{tabular}{lcc}
	\toprule[0.1em]
	Method & Detection (IoU)$\uparrow$ & Inst. Segmentation (IoU)$\uparrow$ \\ 
	\midrule
	SAMP~\cite{engelmann2017samp} & 90.39 & 89.58 \\
	NVDiffRec~\cite{munkberg2022extracting} & 85.26 & 85.88 \\
	NeRF++~\cite{zhang2020nerf++} & 92.81 & 93.22 \\
	NeuS~\cite{wang2021neus} & 93.97 & 94.22 \\
	\rowcolor{grey} Ours & \textbf{94.82} & \textbf{95.48} \\
	\bottomrule
	\end{tabular}
    \end{center}
    \cuthalftablecaptionup
    \caption{Evaluate downstream perception tasks on camera simulation}
    \cuthalftablecaptiondown
    \vspace{-10pt}
    \label{tab:downstream_evaluation}
\end{table}

\cutsectionup
\section{Conclusion}
\cutsectiondown
In this paper, we propose \name, a novel approach for 3D object reconstruction and novel view synthesis from in-the-wild camera and LiDAR data.
\name represents the object geometry as a neural SDF, and the appearance with a physics-based reflectance model.
With this decomposed representation, we can realistically and efficiently render assets at novel views.
We demonstrated that \name assets can be inserted into new scenarios, generate realistic multi-sensor data, and can be used to evaluate autonomy perception, enabling scalable and diverse simulation for self-driving.
Future work involves explicitly modelling scene lighting~\cite{nerfactor,wang2022neural}, large-scale and dynamic scene~\cite{tancik2022blocknerf,rematas2021urban,ost2021neural}, efficient training and real-time rendering~\cite{garbin2021fastnerf,muller2022instant}, and dealing with inaccurate sensor poses~\cite{lin2021barf,yen2021inerf}.

\cutsectionup
\section{Acknowledgements}
\cutsectiondown
We are grateful to Anqi Joyce Yang, Ioan-Andrei Barsan and Wei-Chiu Ma for profound discussion, constructive feedback and proofreading.
We also thank the Waabi team for their valuable assistance and support.

\bibliographystyle{IEEEtran}
\bibliography{egbib}

\begin{thebibliography}{10}
\providecommand{\url}[1]{#1}
\csname url@rmstyle\endcsname
\providecommand{\newblock}{\relax}
\providecommand{\bibinfo}[2]{#2}
\providecommand\BIBentrySTDinterwordspacing{\spaceskip=0pt\relax}
\providecommand\BIBentryALTinterwordstretchfactor{4}
\providecommand\BIBentryALTinterwordspacing{\spaceskip=\fontdimen2\font plus
\BIBentryALTinterwordstretchfactor\fontdimen3\font minus
  \fontdimen4\font\relax}
\providecommand\BIBforeignlanguage[2]{{%
\expandafter\ifx\csname l@#1\endcsname\relax
\typeout{** WARNING: IEEEtran.bst: No hyphenation pattern has been}%
\typeout{** loaded for the language `#1'. Using the pattern for}%
\typeout{** the default language instead.}%
\else
\language=\csname l@#1\endcsname
\fi
#2}}

\bibitem{dosovitskiy2017carla}
A.~Dosovitskiy, G.~Ros, F.~Codevilla, A.~Lopez, and V.~Koltun, ``{CARLA}: An
  open urban driving simulator,'' in \emph{{CoRL}}, 2017.

\bibitem{shah2018airsim}
S.~Shah, D.~Dey, C.~Lovett, and A.~Kapoor, ``{AirSim}: High-fidelity visual and
  physical simulation for autonomous vehicles,'' in \emph{Field and service
  robotics}, 2018.

\bibitem{mildenhall2020nerf}
B.~Mildenhall, P.~P. Srinivasan, M.~Tancik, J.~T. Barron, R.~Ramamoorthi, and
  R.~Ng, ``{NeRF}: Representing scenes as neural radiance fields for view
  synthesis,'' in \emph{ECCV}, 2020.

\bibitem{kajiya1984ray}
J.~T. Kajiya and B.~P. Von~Herzen, ``Ray tracing volume densities,'' 1984.

\bibitem{zhang2020nerf++}
K.~Zhang, G.~Riegler, N.~Snavely, and V.~Koltun, ``{NeRF}++: Analyzing and
  improving neural radiance fields,'' \emph{arXiv}, 2020.

\bibitem{wang2021neus}
P.~Wang, L.~Liu, Y.~Liu, C.~Theobalt, T.~Komura, and W.~Wang, ``Neus: Learning
  neural implicit surfaces by volume rendering for multi-view reconstruction,''
  2021.

\bibitem{Oechsle2021ICCV}
M.~Oechsle, S.~Peng, and A.~Geiger, ``{UNISURF}: Unifying neural implicit
  surfaces and radiance fields for multi-view reconstruction,'' in \emph{ICCV},
  2021.

\bibitem{jain2021putting}
A.~Jain, M.~Tancik, and P.~Abbeel, ``Putting nerf on a diet: Semantically
  consistent few-shot view synthesis,'' in \emph{ICCV}, 2021.

\bibitem{barron2021mipnerf360}
J.~T. Barron, B.~Mildenhall, D.~Verbin, P.~P. Srinivasan, and P.~Hedman,
  ``Mip-nerf 360: Unbounded anti-aliased neural radiance fields,''
  \emph{arXiv}, 2021.

\bibitem{hoffman2018cycada}
J.~Hoffman, E.~Tzeng, T.~Park, J.-Y. Zhu, P.~Isola, K.~Saenko, A.~Efros, and
  T.~Darrell, ``{CyCADA}: Cycle-consistent adversarial domain adaptation,'' in
  \emph{ICML}, 2018.

\bibitem{manivasagam2020lidarsim}
S.~Manivasagam, S.~Wang, K.~Wong, W.~Zeng, M.~Sazanovich, S.~Tan, B.~Yang,
  W.-C. Ma, and R.~Urtasun, ``{LiDARsim}: Realistic lidar simulation by
  leveraging the real world,'' in \emph{CVPR}, 2020.

\bibitem{yang2020surfelgan}
Z.~Yang, Y.~Chai, D.~Anguelov, Y.~Zhou, P.~Sun, D.~Erhan, S.~Rafferty, and
  H.~Kretzschmar, ``Surfelgan: Synthesizing realistic sensor data for
  autonomous driving,'' in \emph{CVPR}, 2020.

\bibitem{engelmann2017samp}
F.~Engelmann, J.~St{\"u}ckler, and B.~Leibe, ``Samp: shape and motion priors
  for 4d vehicle reconstruction,'' in \emph{WACV}.\hskip 1em plus 0.5em minus
  0.4em\relax IEEE, 2017, pp. 400--408.

\bibitem{lu2020permo}
F.~Lu, Z.~Liu, X.~Song, D.~Zhou, W.~Li, H.~Miao, M.~Liao, L.~Zhang, B.~Zhou,
  R.~Yang, \emph{et~al.}, ``{PerMO}: Perceiving more at once from a single
  image for autonomous driving,'' \emph{arXiv}, 2020.

\bibitem{yang2020recovering}
Z.~Yang, S.~Manivasagam, M.~Liang, B.~Yang, W.-C. Ma, and R.~Urtasun,
  ``Recovering and simulating pedestrians in the wild,'' in \emph{CoRL}, 2020.

\bibitem{duggal2021secrets}
S.~Duggal, Z.~Wang, W.-C. Ma, S.~Manivasagam, J.~Liang, S.~Wang, and
  R.~Urtasun, ``Secrets of 3d implicit object shape reconstruction in the
  wild,'' \emph{arXiv}, 2021.

\bibitem{zhao2020vehicle}
X.~Zhao, Z.~Zheng, C.~Ji, Z.~Liu, S.~Lin, T.~Yu, J.~Suo, and Y.~Liu, ``Vehicle
  reconstruction and texture estimation using deep implicit semantic template
  mapping,'' \emph{arXiv}, 2020.

\bibitem{chen2021geosim}
Y.~Chen, F.~Rong, S.~Duggal, S.~Wang, X.~Yan, S.~Manivasagam, S.~Xue, E.~Yumer,
  and R.~Urtasun, ``Geosim: Realistic video simulation via geometry-aware
  composition for self-driving,'' in \emph{CVPR}, 2021.

\bibitem{kim2021drivegan}
S.~W. Kim, J.~Philion, A.~Torralba, and S.~Fidler, ``{DriveGAN}: Towards a
  controllable high-quality neural simulation,'' in \emph{CVPR}, 2021.

\bibitem{park2019deepsdf}
J.~J. Park, P.~Florence, J.~Straub, R.~Newcombe, and S.~Lovegrove, ``{DeepSDF}:
  Learning continuous signed distance functions for shape representation,'' in
  \emph{CVPR}, 2019.

\bibitem{chen2019learning}
Z.~Chen and H.~Zhang, ``Learning implicit fields for generative shape
  modeling,'' in \emph{CVPR}, 2019.

\bibitem{yang2021s3}
Z.~Yang, S.~Wang, S.~Manivasagam, Z.~Huang, W.-C. Ma, X.~Yan, E.~Yumer, and
  R.~Urtasun, ``S3: Neural shape, skeleton, and skinning fields for 3d human
  modeling,'' in \emph{CVPR}, 2021, pp. 13\,284--13\,293.

\bibitem{mescheder2019occupancy}
L.~Mescheder, M.~Oechsle, M.~Niemeyer, S.~Nowozin, and A.~Geiger, ``Occupancy
  networks: Learning 3d reconstruction in function space,'' in \emph{CVPR},
  2019.

\bibitem{atzmon2020sal}
M.~Atzmon and Y.~Lipman, ``{SAL}: Sign agnostic learning of shapes from raw
  data,'' in \emph{CVPR}, 2020.

\bibitem{davies2020effectiveness}
T.~Davies, D.~Nowrouzezahrai, and A.~Jacobson, ``On the effectiveness of
  weight-encoded neural implicit 3d shapes,'' \emph{arXiv}, 2020.

\bibitem{sitzmann2021awesome}
V.~Sitzmann, ``Awesome implicit representations - a curated list of resources
  on implicit neural representations.''

\bibitem{sitzmann2019scene}
V.~Sitzmann, M.~Zollh{\"o}fer, and G.~Wetzstein, ``Scene representation
  networks: Continuous 3d-structure-aware neural scene representations,'' 2019.

\bibitem{sitzmann2020implicit}
V.~Sitzmann, J.~Martel, A.~Bergman, D.~Lindell, and G.~Wetzstein, ``Implicit
  neural representations with periodic activation functions,''
  \emph{{NeurIPS}}, 2020.

\bibitem{niemeyer2020differentiable}
M.~Niemeyer, L.~Mescheder, M.~Oechsle, and A.~Geiger, ``Differentiable
  volumetric rendering: Learning implicit 3d representations without 3d
  supervision,'' in \emph{CVPR}, 2020.

\bibitem{yariv2021volume}
L.~Yariv, J.~Gu, Y.~Kasten, and Y.~Lipman, ``Volume rendering of neural
  implicit surfaces,'' in \emph{NeurIPS}, 2021.

\bibitem{azinovic2021neural}
D.~Azinovi{\'c}, R.~Martin-Brualla, D.~B. Goldman, M.~Nie{\ss}ner, and
  J.~Thies, ``Neural {RGB-D} surface reconstruction,'' \emph{arXiv}, 2021.

\bibitem{insafutdinov2022snes}
E.~Insafutdinov, D.~Campbell, J.~F. Henriques, and A.~Vedaldi, ``Snes: Learning
  probably symmetric neural surfaces from incomplete data,'' in \emph{ECCV},
  2022.

\bibitem{deng2021depth}
K.~Deng, A.~Liu, J.-Y. Zhu, and D.~Ramanan, ``Depth-supervised nerf: Fewer
  views and faster training for free,'' \emph{arXiv}, 2021.

\bibitem{nerfactor}
X.~Zhang, P.~P. Srinivasan, B.~Deng, P.~Debevec, W.~T. Freeman, and J.~T.
  Barron, ``{NeRFactor}: Neural factorization of shape and reflectance under an
  unknown illumination,'' \emph{arXiv}, 2021.

\bibitem{srinivasan2021nerv}
P.~P. Srinivasan, B.~Deng, X.~Zhang, M.~Tancik, B.~Mildenhall, and J.~T.
  Barron, ``{NeRV}: Neural reflectance and visibility fields for relighting and
  view synthesis,'' in \emph{CVPR}, 2021.

\bibitem{bi2020neural}
S.~Bi, Z.~Xu, P.~Srinivasan, B.~Mildenhall, K.~Sunkavalli, M.~Ha{\v{s}}an,
  Y.~Hold-Geoffroy, D.~Kriegman, and R.~Ramamoorthi, ``Neural reflectance
  fields for appearance acquisition,'' \emph{arXiv}, 2020.

\bibitem{boss2021nerd}
M.~Boss, R.~Braun, V.~Jampani, J.~T. Barron, C.~Liu, and H.~Lensch, ``{NeRD}:
  Neural reflectance decomposition from image collections,'' in \emph{ICCV},
  2021.

\bibitem{zhang2021physg}
K.~Zhang, F.~Luan, Q.~Wang, K.~Bala, and N.~Snavely, ``Physg: Inverse rendering
  with spherical gaussians for physics-based material editing and relighting,''
  in \emph{CVPR}, 2021.

\bibitem{phong1975illumination}
B.~T. Phong, ``Illumination for computer generated pictures,''
  \emph{Communications of the ACM}, 1975.

\bibitem{arvo1995applications}
J.~Arvo, ``Applications of irradiance tensors to the simulation of
  non-lambertian phenomena,'' in \emph{Proceedings of the 22nd annual
  conference on Computer graphics and interactive techniques}, 1995.

\bibitem{ost2021neural}
J.~Ost, F.~Mannan, N.~Thuerey, J.~Knodt, and F.~Heide, ``Neural scene graphs
  for dynamic scenes,'' in \emph{CVPR}, 2021.

\bibitem{niemeyer2021giraffe}
M.~Niemeyer and A.~Geiger, ``{GIRAFFE}: Representing scenes as compositional
  generative neural feature fields,'' in \emph{CVPR}, 2021.

\bibitem{liu2021editing}
S.~Liu, X.~Zhang, Z.~Zhang, R.~Zhang, J.-Y. Zhu, and B.~Russell, ``Editing
  conditional radiance fields,'' \emph{arXiv}, 2021.

\bibitem{yang2021learning}
B.~Yang, Y.~Zhang, Y.~Xu, Y.~Li, H.~Zhou, H.~Bao, G.~Zhang, and Z.~Cui,
  ``Learning object-compositional neural radiance field for editable scene
  rendering,'' in \emph{ICCV}, 2021.

\bibitem{schwarz2020graf}
K.~Schwarz, Y.~Liao, M.~Niemeyer, and A.~Geiger, ``{GRaF}: Generative radiance
  fields for 3d-aware image synthesis,'' 2020.

\bibitem{guo2020object}
M.~Guo, A.~Fathi, J.~Wu, and T.~Funkhouser, ``Object-centric neural scene
  rendering,'' 2021.

\bibitem{chen2021dib}
W.~Chen, J.~Litalien, J.~Gao, Z.~Wang, C.~F. Tsang, S.~Khamis, O.~Litany, and
  S.~Fidler, ``{DIB-R++}: Learning to predict lighting and material with a
  hybrid differentiable renderer,'' \emph{arXiv}, 2021.

\bibitem{zhang2021ners}
J.~Y. Zhang, G.~Yang, S.~Tulsiani, and D.~Ramanan, ``Ners: Neural reflectance
  surfaces for sparse-view 3d reconstruction in the wild,'' in \emph{NeurIPS},
  2021.

\bibitem{barrow1978recovering}
H.~Barrow, J.~Tenenbaum, A.~Hanson, and E.~Riseman, ``Recovering intrinsic
  scene characteristics,'' \emph{Comput. Vis. Syst}, vol.~2, no. 3-26, p.~2,
  1978.

\bibitem{barron2013intrinsic}
J.~T. Barron and J.~Malik, ``Intrinsic scene properties from a single {RGB-D}
  image,'' in \emph{CVPR}, 2013.

\bibitem{ma2018single}
W.-C. Ma, H.~Chu, B.~Zhou, R.~Urtasun, and A.~Torralba, ``Single image
  intrinsic decomposition without a single intrinsic image,'' in \emph{ECCV},
  2018.

\bibitem{barron2021mip}
J.~T. Barron, B.~Mildenhall, M.~Tancik, P.~Hedman, R.~Martin-Brualla, and P.~P.
  Srinivasan, ``{Mip-NeRF}: A multiscale representation for anti-aliasing
  neural radiance fields,'' 2021.

\bibitem{immel1986radiosity}
D.~S. Immel, M.~F. Cohen, and D.~P. Greenberg, ``A radiosity method for
  non-diffuse environments,'' 1986.

\bibitem{kajiya1986rendering}
J.~T. Kajiya, ``The rendering equation,'' in \emph{Proceedings of the 13th
  annual conference on Computer graphics and interactive techniques}, 1986.

\bibitem{walter2007microfacet}
B.~Walter, S.~R. Marschner, H.~Li, and K.~E. Torrance, ``Microfacet models for
  refraction through rough surfaces.'' \emph{Rendering techniques}, vol. 2007,
  p. 18th, 2007.

\bibitem{kirillov2020pointrend}
A.~Kirillov, Y.~Wu, K.~He, and R.~Girshick, ``Pointrend: Image segmentation as
  rendering,'' in \emph{CVPR}, 2020.

\bibitem{tulsiani2018layer}
S.~Tulsiani, R.~Tucker, and N.~Snavely, ``Layer-structured 3d scene inference
  via view synthesis,'' in \emph{ECCV}, 2018, pp. 302--317.

\bibitem{munkberg2022extracting}
J.~Munkberg, J.~Hasselgren, T.~Shen, J.~Gao, W.~Chen, A.~Evans, T.~M{\"u}ller,
  and S.~Fidler, ``Extracting triangular 3d models, materials, and lighting
  from images,'' in \emph{CVPR}, 2022, pp. 8280--8290.

\bibitem{xiao2021pandaset}
P.~Xiao, Z.~Shao, S.~Hao, Z.~Zhang, X.~Chai, J.~Jiao, Z.~Li, J.~Wu, K.~Sun,
  K.~Jiang, \emph{et~al.}, ``Pandaset: Advanced sensor suite dataset for
  autonomous driving,'' in \emph{ITSC}, 2021.

\bibitem{zhang2018perceptual}
R.~Zhang, P.~Isola, A.~A. Efros, E.~Shechtman, and O.~Wang, ``The unreasonable
  effectiveness of deep features as a perceptual metric,'' in \emph{CVPR},
  2018.

\bibitem{lorensen1987marching}
W.~E. Lorensen and H.~E. Cline, ``Marching cubes: A high resolution 3d surface
  construction algorithm,'' \emph{ACM siggraph computer graphics}, vol.~21,
  no.~4, pp. 163--169, 1987.

\bibitem{fang2021lidar}
J.~Fang, X.~Zuo, D.~Zhou, S.~Jin, S.~Wang, and L.~Zhang, ``Lidar-aug: A general
  rendering-based augmentation framework for 3d object detection,'' in
  \emph{CVPR}, 2021.

\bibitem{wang2021advsim}
J.~Wang, A.~Pun, J.~Tu, S.~Manivasagam, A.~Sadat, S.~Casas, M.~Ren, and
  R.~Urtasun, ``Advsim: Generating safety-critical scenarios for self-driving
  vehicles,'' in \emph{CVPR}, 2021.

\bibitem{wang2022neural}
Z.~Wang, W.~Chen, D.~Acuna, J.~Kautz, and S.~Fidler, ``Neural light field
  estimation for street scenes with differentiable virtual object insertion,''
  in \emph{ECCV}.\hskip 1em plus 0.5em minus 0.4em\relax Springer, 2022, pp.
  380--397.

\bibitem{tancik2022blocknerf}
M.~Tancik, V.~Casser, X.~Yan, S.~Pradhan, B.~Mildenhall, P.~Srinivasan, J.~T.
  Barron, and H.~Kretzschmar, ``{Block-NeRF}: Scalable large scene neural view
  synthesis,'' \emph{arXiv}, 2022.

\bibitem{rematas2021urban}
K.~Rematas, A.~Liu, P.~P. Srinivasan, J.~T. Barron, A.~Tagliasacchi,
  T.~Funkhouser, and V.~Ferrari, ``Urban radiance fields,'' \emph{arXiv
  preprint arXiv:2111.14643}, 2021.

\bibitem{garbin2021fastnerf}
S.~J. Garbin, M.~Kowalski, M.~Johnson, J.~Shotton, and J.~Valentin, ``Fastnerf:
  High-fidelity neural rendering at 200fps,'' in \emph{ICCV}, 2021, pp.
  14\,346--14\,355.

\bibitem{muller2022instant}
T.~M{\"u}ller, A.~Evans, C.~Schied, and A.~Keller, ``Instant neural graphics
  primitives with a multiresolution hash encoding,'' \emph{ACM Transactions on
  Graphics (ToG)}, vol.~41, no.~4, pp. 1--15, 2022.

\bibitem{lin2021barf}
C.-H. Lin, W.-C. Ma, A.~Torralba, and S.~Lucey, ``Barf: Bundle-adjusting neural
  radiance fields,'' in \emph{ICCV}, 2021, pp. 5741--5751.

\bibitem{yen2021inerf}
L.~Yen-Chen, P.~Florence, J.~T. Barron, A.~Rodriguez, P.~Isola, and T.-Y. Lin,
  ``inerf: Inverting neural radiance fields for pose estimation,'' in
  \emph{IROS}.\hskip 1em plus 0.5em minus 0.4em\relax IEEE, 2021, pp.
  1323--1330.

\bibitem{yariv2020multiview}
L.~Yariv, Y.~Kasten, D.~Moran, M.~Galun, M.~Atzmon, R.~Basri, and Y.~Lipman,
  ``Multiview neural surface reconstruction by disentangling geometry and
  appearance,'' in \emph{NeurIPS}, 2020.

\end{thebibliography}

\clearpage

\onecolumn

\setcounter{section}{0}
\renewcommand{\thesection}{S\arabic{section}}
\setcounter{table}{0}
\setcounter{figure}{0}
\section*{Supplementary}

In the supplementary material, we provide implementation details about our method and the experiments, additional qualitative visualizations, and limitations of our method. 
We first describe the details of our dataset in Sec.~\ref{sec:dataset_details}.
Then we describe the implementation details of our model in Sec.~\ref{sec:implementation_details}.
Next, we show additional details of the baseline models in Sec.~\ref{sec:baseline_details}.
After that, we showcase additional visualizations and results on downstream applications in Sec.~\ref{sec:additional_results}.
Finally, we analyze the limitations of our model in Sec.~\ref{sec:limitation}.

\begin{figure*}[h]
\begin{center}
	\includegraphics[width=0.98\textwidth]{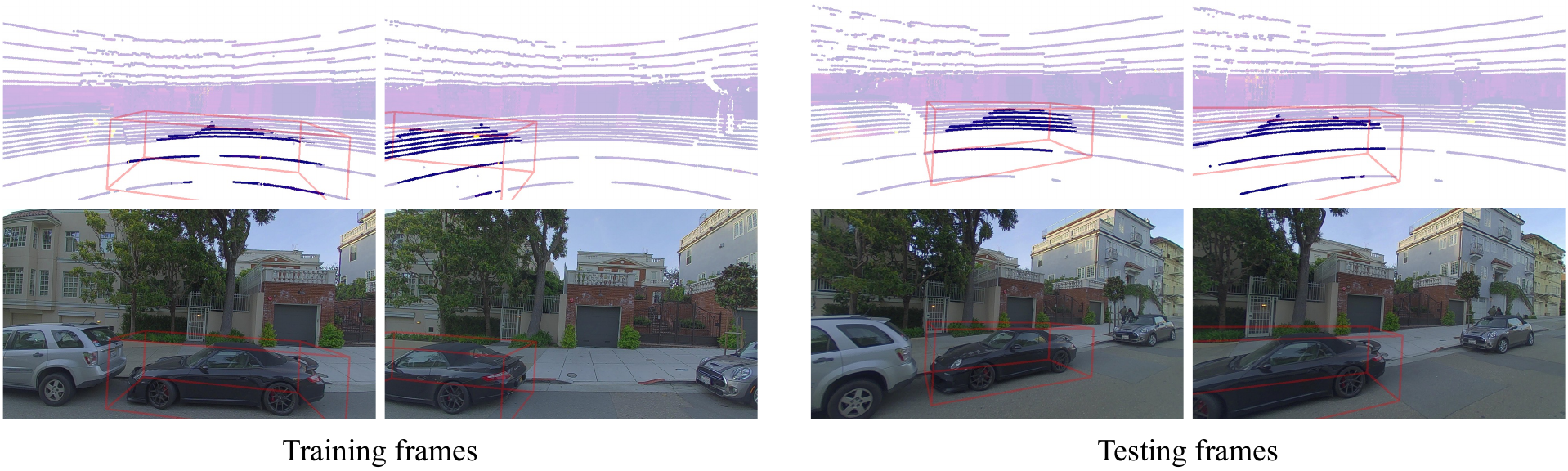}
\end{center}
\caption{Example training and testing frames of the PandaVehicle dataset.}
\label{fig:dataset}
\end{figure*}

\section{Dataset and Experimental Settings}
\label{sec:dataset_details}
The PandaVehicle dataset contains 10 vehicles curated from PandaSet~\cite{xiao2021pandaset} with diverse shape and appearance under complex illumination and occlusion. 
The data was captured by a self-driving vehicle platform equipped with six cameras (front, front-left camera, left, front-right, right and back cameras) and two LiDARs (a $360^\circ$ mechanical spinning LiDAR and a forward-facing LiDAR).
All the sensors are calibrated.
Each asset is captured  when the self-driving vehicle (SDV) passes by. 
We employ the left camera for training and the front-left camera for evaluation, and we also use the $360^\circ$ mechanical spinning LiDAR to train the model. 
Each asset has on average $\sim$24 views for training.
Please refer to Fig.~\ref{fig:dataset} and supplementary video for example frames.
Since we focus on foreground vehicles, we use an off-the-shelf~\cite{kirillov2020pointrend} algorithm to estimate the segmentation mask for each frame in the camera video.
In addition to using the inferred segmentation masks as supervision for \name, they are also used for evaluation to filter foreground pixels in the quantitative comparison.
Please see Table~\ref{tab:dataset} for detailed information for all 10 selected vehicles.

\begin{table*}[ht]
	\centering
	\resizebox{\textwidth}{!}{
		\begin{tabular}{cc c c c}
			\toprule
			\textbf{Actor UUID} & \textbf{Log ID} & \textbf{Train frames (Left camera)} & \textbf{Test frames (Front-left camera)}\\
			\midrule
			1be68ce6-68c5-467f-abb1-fa5e03d1db7a & 053 & 33-36, 40-49 & 25-41\\
			\midrule
			1d79eded-2fb0-4f89-ba35-323926f45ade & 139 & 46-63 & 44-55\\
			\midrule
			2160d735-3fda-49f8-9bd9-e2cba3b51faa & 038 & 34-47 & 27-41\\
			\midrule
			2ee4d8f8-af0a-48f3-bb6c-ed479a7829e7 & 039 & 47-67 & 28, 31-59\\
			\midrule
			526e2f5e-e294-415c-aad6-578d27921465 & 030 & 38-78 & 35-55\\
			\midrule
			56e10a51-35ed-43b0-837c-cea8aff216cc & 139 & 26-52 & 25-46\\
			\midrule
			5ce5fb69-038d-4f82-8c64-90b73c6f6681 & 030 & 17-62 & 0-45\\
			\midrule
			94c06b25-d17a-4ee7-a2df-7faa619bee89 & 035 & 49-58, 60-61 & 47-51\\
			\midrule
			ba222d39-2f13-4849-8ff4-91e247d5cedf & 120 & 12-37 & 0-25\\
			\midrule
			f7bd1486-1fbe-4f33-ba28-f00dae3e0298 & 139 & 57-77 & 54-69\\
			\bottomrule
		\end{tabular}
	}
	\caption{Asset ID, Log ID and camera image ID for the 10 selected assets from PandaSet~\cite{xiao2021pandaset}.}
	\label{tab:dataset}
\end{table*}

\begin{figure*}[t]
\begin{center}
	\includegraphics[width=0.98\textwidth]{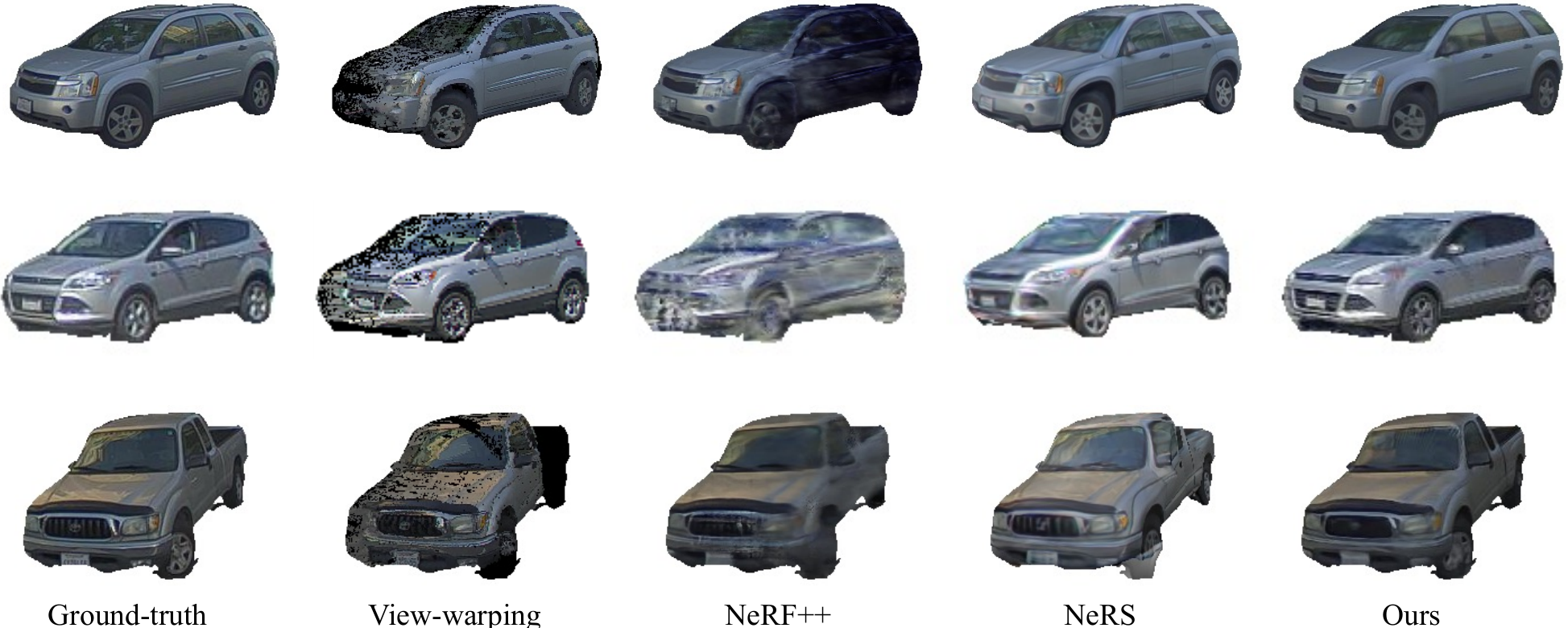}
\end{center}
\caption{Novel view synthesis results on challenging data.}
\label{fig:novel_view}
\end{figure*}

\begin{figure*}[t]
\begin{center}
    \includegraphics[width=0.98\textwidth]{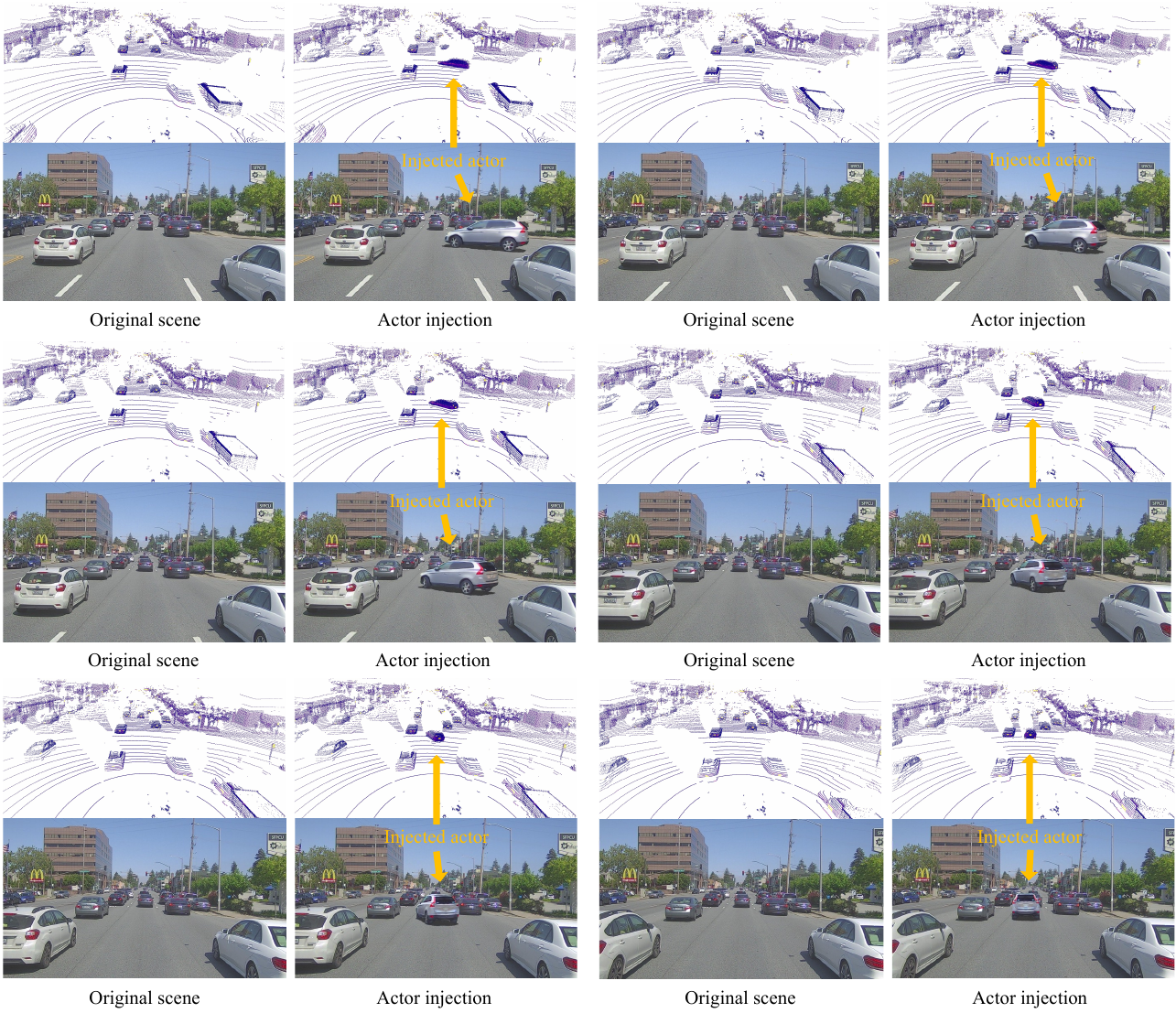}
\end{center}
\caption{
We can generate consistent multi-sensor simulation using our reconstructed assets. The reconstructed asset performs a right turn and merges into the main road. We visualized LiDAR and camera data for sampled frames with/without the added actor. 
Time increases going from left to right and from top to bottom.
}
\label{fig:simulation_right_turn}
\end{figure*}

\begin{figure*}[t]
\begin{center}
    \includegraphics[width=0.98\textwidth]{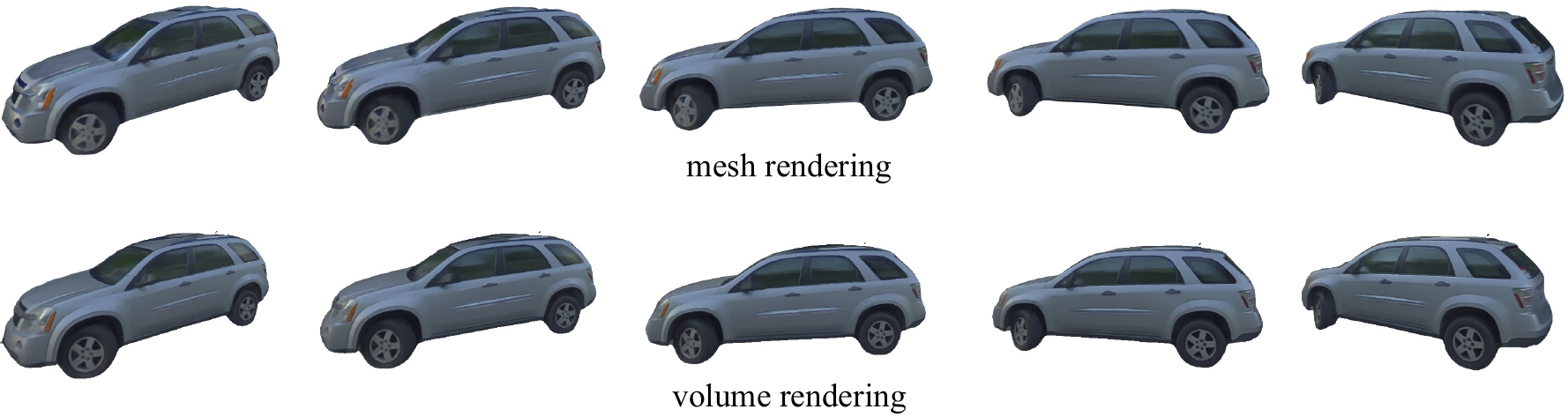}
\end{center}
\caption{Visual comparison of mesh rendering and volume rendering. From left to right we show the rotation of the asset from different viewpoints.}
\label{fig:efficient_rendering}
\end{figure*}

\begin{figure*}[t]
\begin{center}
    \includegraphics[width=0.98\textwidth]{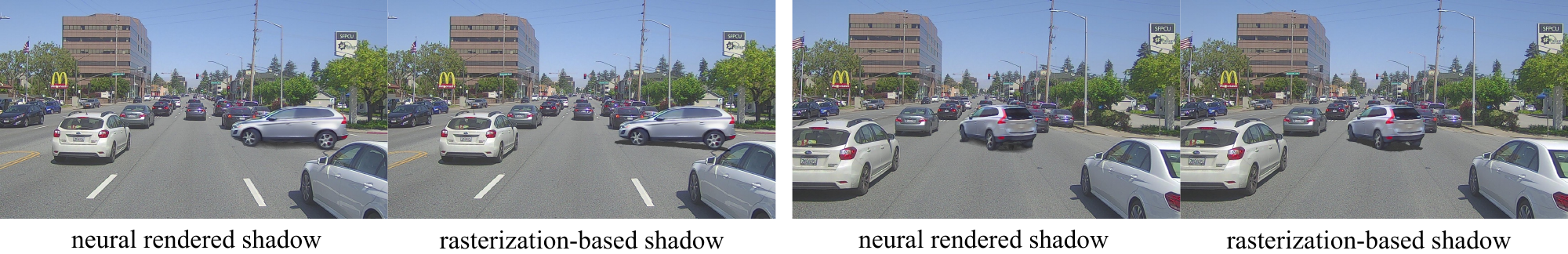}
\end{center}
\caption{
Visualization of shadows generated by different approaches.
We render shadows for the added actor at two different poses.
For neural rendered shadow (left image in each pair), we extract the shadow weight from \name's rendered results with the ground. 
For rasterization-based shadow (right image in each pair), we use a rasterization engine to generate the shadow based on the geometry of the inserted actor, assuming a top-down light.
}
\label{fig:shadow}
\end{figure*}

\begin{figure*}[t]
\begin{center}
	\includegraphics[width=0.98\textwidth]{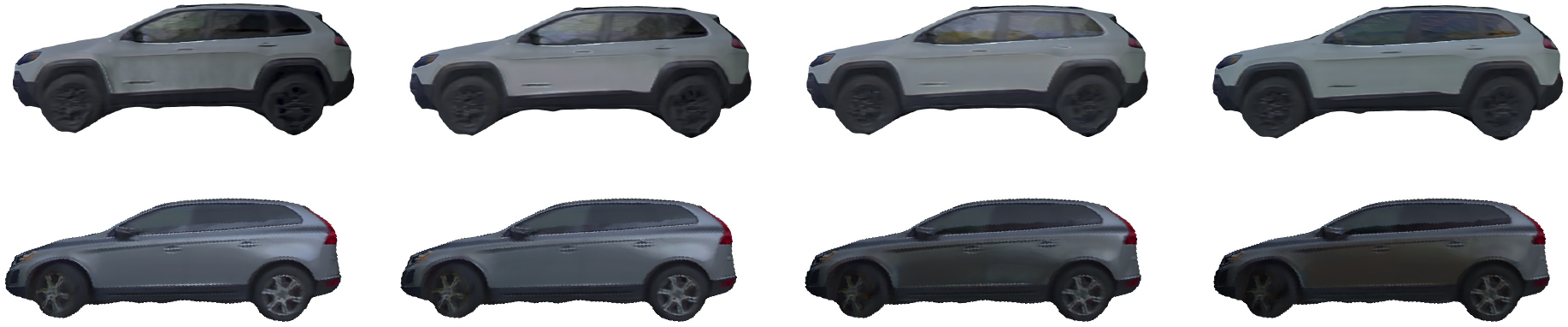}
\end{center}
\caption{
Relighting results when rotating the environmental lighting map $\mathcal E$. 
Top: Asset example 1. Bottom: Asset example 2. From left to right we show the change in environmental lighting rotation.
} 
\label{fig:relighting}
\end{figure*}

\section{Implementation Details}
\label{sec:implementation_details}

Our network architecture is similar to IDR~\cite{yariv2020multiview} and consists of three MLPs to encode the SDF, albedo, and material shininess.
The SDF network $f_\text{SDF}$ is a 8-layer MLP with a hidden size of 256 and Softplus activation.
A skip connection is used to connect the input with the output of the fourth layer in the SDF MLP.
We initialize the SDF MLP so it produces an approximate SDF of a sphere~\cite{atzmon2020sal}.
We initialize the learnable parameter $\beta = 20$ in the sigmoid-like function for conversion from SDF to $\alpha$ (Eqn. (3) in the main paper).
The reflectance network $f_\text{reflectance}$ are implemented as 4-layer MLPs with a hidden size of 256 and ReLU activation. 
Both MLPs take as input the point location $\mathbf x$, normal $\mathbf n$ and the feature output of the last layer $f_\text{SDF}$.
We use the Adam optimizer with learning rate of $5\mathrm{e}{-4}$.  The coefficients used for computing running averages are left at default values of $\beta_1 = 0.9$, $\beta_2 = 0.999$, and $\epsilon = 1\mathrm{e}{-8}$.
During training, the loss weights are $\lambda_\text{lidar} = 0.1$, $\lambda_\text{mask} = 0.1$, $\lambda_\text{Eik} = 0.1$ for SDF regularizer and $\lambda_\text{sym} = 1.0$ for structural symmetry prior, respectively.

\subsubsection{Background Model}
Outdoor scenes contain backgrounds (buildings, trees, sky, \etc) that are arbitrarily far away, which leads to resolution issues in volume rendering (Eqn. (1) in the main paper).
NeRF~\cite{mildenhall2020nerf} addresses this issue by utilizing Normalized Device Coordinates (NDC) parametrization, but it cannot cover the space outside the reference view's frustrum.
To address this limitation, NeRF++~\cite{zhang2020nerf++} propose to use an inverted sphere parameterization~\cite{zhang2020nerf++} to participate space into an inner sphere volume and outer volume, where the foreground objects and all the cameras are normalized inside the inner sphere volume.
This works well for $360^{\circ}$ captures centered on close objects.
However, this can be problematic when the foreground objects are far from the sensors or the sensors move along a long trajectory path. 
This formulation would require a large inner sphere volume to cover the space of foreground objects and all sensors, and this hurts the sampling resolution.
In our scene representation, we assume the rendered ray $r(t)$ intersects with the object's region of interests (acquired from an annotation or an automatic detection output) at $t_\text{near}$ and $t_\text{far}$. 
We divide the traversed space into foreground ($\{t_\text{near}<t<t_\text{far} \}$) and background.
To compute the background scene radiance, we sample the ray's intersections with Multiple-Sphere Images (MPI) surrounding the object of interest. 
We generate the radii for MSI by linearly interpolating inverse depths.

\section{Baseline Details}
\label{sec:baseline_details}
We compare our model with the state-of-the-art baselines: the volume rendering method NeRF++~\cite{zhang2020nerf++}, NeuS~\cite{wang2021neus} and the inverse graphics model NeRS~\cite{zhang2021ners}, NVDiffRec~\cite{munkberg2022extracting}.
We choose these baselines as they model reflectance and work well in our outdoor setting.
We also compare against geometry-based method LiDAR-guided view-warping~\cite{tulsiani2018layer} and SAMP~\cite{engelmann2017samp}.
Next, we next provide the implementation details of these baseline models.

\subsection{Volume Rendering-based Baselines}

\subsubsection{NeRF++}
For NeRF++~\cite{zhang2020nerf++}, we do scene normalization to move the camera's trajectory within the unit sphere. We adopt the same hyperparameters from the original code\footnote{\url{https://github.com/Kai-46/nerfplusplus}}, except we train $100k$ iterations because the PandaVehicle dataset has sparser views and the model converges faster.

\subsubsection{NeuS}
For NeuS~\cite{wang2021neus}, we follow the official code repository\footnote{\url{https://github.com/Totoro97/NeuS}} to perform scene normalization to make the asset’s region of interest fall inside the unit sphere, and model the background by a separate model similar to NeRF++~\cite{zhang2020nerf++}.
We train each asset for 200k iterations.

\subsection{Inverse Graphics-based Baselines}
\subsubsection{NeRS}
For NeRS\footnote{\url{https://github.com/jasonyzhang/ners}} baseline, we downscale the input images $4 \times$ (in contrast to $2 \times$ in NeRSDF and NeRF++) due to GPU memory limitations.
We initialized the cuboid template with the assets' coarse 3D dimensions and set the level of unit ico-sphere as 6.
As the precise camera poses are given, we employ a three-stage training process: sequentially optimizing the shape, texture, and illumination parameters. 
To ensure better visual quality and semantic metrics, we increased the weights of the chamfer loss and perceptual loss to 0.04 and 1.0, respectively. 
We also removed off-screen loss as not all input views contain the complete vehicle shapes.
Moreover, we increased the training iterations on the three stages to $3k, 12k, 3k$ since more input views are provided in the PandaVehicle dataset and it takes longer to converge.
To reconstruct missing parts and obtain better evaluation results on extrapolated views, we also applied symmetry constraints to the deformed textured meshes along the heading axis.
The learning rate of Adam optimizer in the three stages is $1e^{-4}$.

\subsubsection{NVDiffRec}
NVDiffRec~\cite{munkberg2022extracting} is an efficient differentiable rendering-based 3D reconstruction approach that combines differentiable marching tetrahedrons and split-sum environment lighting. 
It achieves SotA performance on a wide variety of synthetic datasets with dense camera views. 
We follow the official code repository\footnote{\url{https://github.com/NVlabs/nvdiffrec}} and set the tetrahedron grid resolution as 64 and the mesh scale as 5.0 (real vehicle scale). 
The model is trained for 5k iterations (batch size 8) with a learning rate exponentially decayed from 0.03 to 0.003.

\subsection{Geometry-based Baselines}
\subsubsection{LiDAR-guided View-warping}
We first aggregate LiDAR points with the Iterative Closest Points (ICP) algorithm to create a surfel representation for the asset.
Given a testing viewpoint, we render the surfel asset to this viewpoint and generate the corresponding depth map. 
Using the rendered depth map, the source camera image and the camera calibration, we generate the object's texture using the inverse warping operation as in~\cite{chen2021geosim}. 
To choose which source image to warp to the target frame, we warp all available source images to the target view and choose the one with the highest overlap with the rendered surfels.

\subsubsection{SAMP}
For SAMP~\cite{engelmann2017samp}, we first processe the CAD library and make each mesh watertight and simplified, we compute volumetric SDFs for each vehicle in metric space (volume dimension 100 × 100 × 100).
Following~\cite{engelmann2017samp}, we apply PCA on the SDF volumes and set the embedding dimension as 25.
In the inference time, we jointly optimize the shape latent code, a scaling factor on the SDF (handle difference shapes) and relative vehicle pose (handle rotation, translation) to fit the LiDAR points. 
We adopt a L1 loss on the SDF value and a total variation loss on the scale factor to penalize abrupt local SDF changes. 
The weights of data and regularization terms are $1$ and $0.1$. 
We use the Adam optimizer with a learning rate of $0.01$.
We use marching cubes to extract the mesh from the optimized SDF volume.
We then use a differentiable renderer to optimize the 2D UV texture for each asset.

\begin{table}[t]
    \begin{center}
    \begin{tabular}{ccccc}
	\toprule[0.1em]
    Rendering & FPS $\uparrow$ & PSNR$\uparrow$ & SSIM$\uparrow$ & LPIPS$\downarrow$ \\
    \midrule
    Volume Rendering& 0.03 & \textbf{22.44} & \textbf{0.692} & \textbf{0.202} \\
    Mesh Rendering& \textbf{76.79} & 19.87 & 0.609 & 0.239 \\
	\bottomrule[0.1em]
    \end{tabular}
    \end{center}
    \caption{Comparisons of volume rendering and mesh rendering.}
    \label{tab:efficient_rendering}
\end{table}

\section{Additional Results}
\label{sec:additional_results}
We now provide more results and details of our model on in-the-wild data.

\subsection{Novel View Synthesis}
We show additional novel view synthesis results in Fig.~\ref{fig:novel_view}. 
Since the recorded vehicles are far away from the cameras, we enlarge the images.
Our model captures more fine-grained details and generalizes better.

\subsection{Efficient Rendering}
We can also render the baked asset efficiently using off-the shelf rasterization engines.
Mesh rendering is on average three orders of magnitude faster than volume rendering (76.79 frames per second (FPS) vs 0.03 FPS for a 960 x 540 image resolution) and still provides good visual quality.
Table~\ref{tab:efficient_rendering} shows an image metrics comparison between the volume rendering result and mesh rendering result. 
The quantitative metrics show that the performance is competitive with other neural volume rendering baseline methods, while being much faster to render.
Fig.~\ref{fig:efficient_rendering} shows a visual comparison between volume rendering and mesh rendering. They look close to indistinguishable.

\subsection{Sensor Simulation}
Importantly, we can use our reconstructed assets to generate consistent multi-sensor simulation for self-driving.
In Fig.~\ref{fig:simulation_right_turn}, the injected NeRSDF vehicle asset performs a right turn and merges into the main road.
We render the image segment using the \name baked asset and a rasterization engine.
Additionally, we render the asset's shadow to ensure the inserted actor region looks realistic. 
We experiment with two approaches. 
In the ``neural rendered'' approach, we extract the shadow weight from \name's rendered results with the ground, and apply it to the background in the target scene. 
In the ``rasterization-based approach,'' we use a rasterization engine and performed shadow mapping to generate the shadow mask for the inserted actor, assuming a top-down light (Fig.~\ref{fig:simulation_right_turn}).
Finally, a neural network is applied to blend the rendered actor and shadow with the background~\cite{chen2021geosim}.
Fig.~\ref{fig:shadow} shows a comparison between shadows generated by the neural-rendered approach and the rasterization-based approach.
To render the simulated LiDAR we use the asset's mesh geometry and perform raycasting according to the LiDAR sensor's extrinsics and intrinsics. We then convert the simulated LiDAR point cloud and real point cloud into a spherical depth-image representation and merge the point clouds to ensure realistic occlusion and LiDAR shadows~\cite{wang2021advsim,fang2021lidar}. 
By using the same baked asset for both camera and LiDAR simulation, we ensure that the simulated data for both sensors match.

\subsubsection{Control over appearance and illumination}
We rotate the environmental lighting map $\mathcal E$ and show the rendering results under novel lighting conditions for two different assets in Fig.~\ref{fig:relighting}.
Please see our video for additional visualizations. With the asset's texture and material properties constant, we can generate realistic variations of the asset views.

\section{Limitations}
\label{sec:limitation}
Our model has difficulties reconstructing certain glass materials that reflect camera rays while LiDAR rays penetrate.
This inconsistency of the sensor observations will cause artifacts on the surface. 
Additionally, our Phong-based reflectance model cannot handle complex reflection and refraction well, or infer shadows casted on the observed view. Given sparse sensor data, we plan to explore how to leverage more complex reflectance and lighting models that better handle these effects in future work. 
Our model also require good camera and LiDAR calibration to ensure alignment of observations across sensors.

\end{document}